\documentclass[letterpaper]{article} 

\usepackage{aaai2027}  

\usepackage[hyphens]{url}  
\usepackage{graphicx} 
\usepackage{natbib}  
\usepackage{caption} 
\usepackage{algorithm}
\usepackage{algpseudocode}
\usepackage{algorithmicx}
\usepackage{newfloat}
\usepackage{listings}
\DeclareCaptionStyle{ruled}{labelfont=normalfont,labelsep=colon,strut=off} 
\floatstyle{ruled}
\newfloat{listing}{tb}{lst}{}
\floatname{listing}{Listing}

\usepackage{booktabs}

\usepackage{microtype}
\usepackage{subcaption}
\usepackage{booktabs}
\usepackage{makecell}
\usepackage{multirow}

\usepackage{amsmath}
\usepackage{amssymb}
\usepackage{mathtools}
\usepackage{amsthm}

\usepackage[capitalize,noabbrev]{cleveref}
\usepackage[textsize=tiny]{todonotes}

\theoremstyle{plain}

\theoremstyle{definition}

\theoremstyle{remark}

\newcommand{\best}[1]{\textcolor{red}{\textbf{#1}}}
\newcommand{\second}[1]{\textcolor{blue}{\underline{#1}}}
\usepackage{tikz}
\usepackage{bm}
\usetikzlibrary{arrows.meta, positioning, calc, backgrounds, fit, shapes.geometric, patterns}

\definecolor{structureblue}{RGB}{37, 99, 235}
\definecolor{structurebg}{RGB}{239, 246, 255}
\definecolor{driftred}{RGB}{220, 38, 38}
\definecolor{driftbg}{RGB}{254, 242, 242}
\definecolor{amberaccent}{RGB}{217, 119, 6}
\definecolor{amberbg}{RGB}{254, 243, 199}
\definecolor{slateborder}{RGB}{148, 163, 184}
\definecolor{panelbg}{RGB}{248, 250, 252}

\definecolor{headerblue}{RGB}{198,211,239}
\definecolor{bodypink}{RGB}{229,211,221}
\definecolor{savegreen}{RGB}{0,130,70}
\usepackage[table]{xcolor}

\title{When Denoising Hurts: Rethinking the Terminal Step of \\
Diffusion Time Series Forecasters}
\author{
    Dat Nguyen-Cong\textsuperscript{\rm 1},
    Luong Tran\textsuperscript{\rm 1},
    Tung Kieu\textsuperscript{\rm 2}\corresponding
}

\affiliations{
    \textsuperscript{\rm 1}FPT Software AI Center, FPT Corporation\\
    \textsuperscript{\rm 2}Department of Computer Science, Aalborg University, Denmark\\
    dat27072002@gmail.com, 
    luongtk@fpt.com,
    tungkvt@cs.aau.dk
}

\begin{document}

\nocopyright
\maketitle

\begin{abstract}
Diffusion models offer a natural way to model uncertainty in time series forecasting, yet their iterative sampling process is often treated as a uniformly beneficial refinement procedure. Our study challenges this view by examining how forecast quality evolves throughout reverse diffusion. 
We find that general temporal structure is often recovered at relatively high noise levels, whereas continued low-noise refinement can introduce statistical drift and degrade the final forecast. 
Our analysis further suggests that this behavior explains why prior methods often favor relatively narrow diffusion architecture and schedule design.
Building on this observation, we propose a label-free global stopping criterion that detects the optimal termination point, eventually speeding up inference and improving predictive accuracy. 
Additionally, since early stopping terminates inference in high-noise regions, we propose a Bernoulli timestep sampler that concentrates training on this region while preserving coverage of the full diffusion process. 
Extensive experiments conducted across eight real-world datasets demonstrate the superior performance of our method compared to existing approaches.

\end{abstract}


\section{Introduction}
\label{sec:introduction}

Time series forecasting remains challenging because observations are often noisy and governed by nonlinear, non-stationary dynamics that evolve over time.
Classical approaches, such as ARMA~\cite{arma} and state-space models~\cite{kvae}, achieve good performance but typically rely on assumptions of linearity, stationarity, or predefined transition dynamics, which may limit their ability to represent complex temporal processes.
Neural autoregressive models~\cite{deepar,lstnet} relax some of these assumptions, yet generate future values sequentially and are therefore susceptible to exposure bias and recursive error accumulation, particularly over long forecasting horizons.
Diffusion models~\cite{ddpm}, on the other hand, offer a compelling alternative by formulating forecasting as conditional generation.
Through iterative denoising, they provide a flexible framework for learning heterogeneous data distributions and explicitly quantifying forecasting uncertainty.

The behavior and design of diffusion models, however, remain insufficiently understood in the context of time series forecasting. Existing approaches have introduced temporal architectures~\cite{timegrad,diffusionts,sssd}, training objectives~\cite{csdi}, and guidance mechanisms~\cite{tsdiff}, yet largely overlook how the structural properties of temporal data interact with the denoising trajectory. 
In contrast, diffusion models in other domains, such as text generation, have been adapted to account for their discrete and sparse characteristics~\cite{dinoiser,difformer}. Only a limited number of time series studies explicitly examine these interactions~\cite{ant,nsdiff}. 
Even so, most existing methods implicitly assume that each reverse-diffusion step monotonically improves the forecast. 
It therefore remains unclear whether all denoising steps are equally beneficial or whether completing the full reverse process is necessary for accurate forecasting.

\begin{figure}[t!]
    \centering

    \begin{subfigure}[t]{0.52\linewidth}
        \centering
        \includegraphics[
            width=\linewidth,
        ]{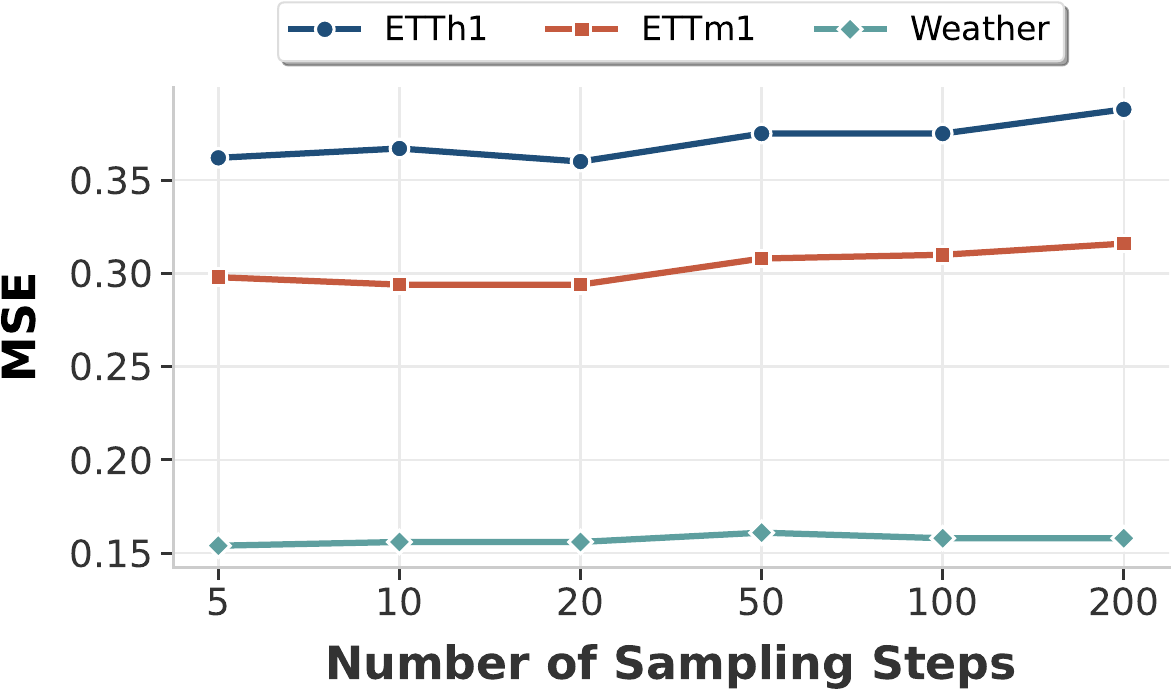}
        \caption{Forecasting performance.}
        \label{fig:nfe_var}
    \end{subfigure}
    \hfill
    \begin{subfigure}[t]{0.47\linewidth}
        \centering
        \includegraphics[
            width=\linewidth,
        ]{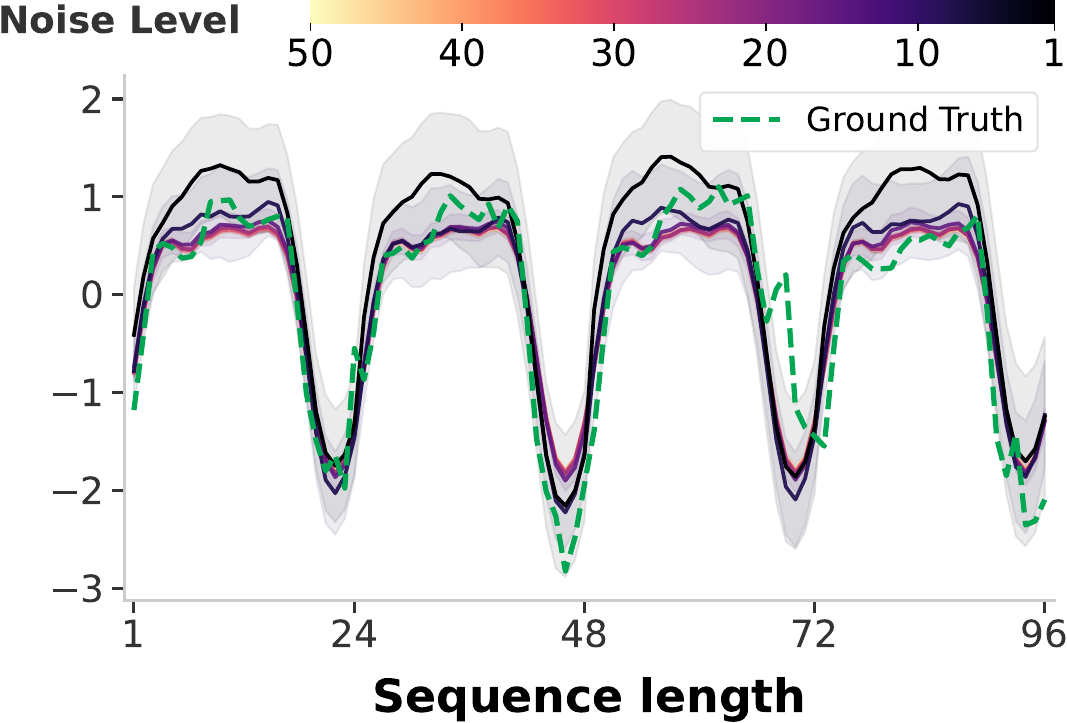}
        \caption{Prediction trajectory.}
        \label{fig:step_predict}
    \end{subfigure}
    \caption{
        (a) Performance is consistent across different numbers of sampling steps, even better with smaller steps.
        (b) During generation, the intermediate predicted samples drift significantly when the noise levels are small.
    }
\end{figure}

\paragraph{Excessive denoising can destabilize and degrade forecasts.} 
As illustrated in Fig.~\ref{fig:nfe_var}, reducing the number of reverse-sampling steps often yields performance comparable to, and in some cases better than, that of substantially more expensive sampling configurations.
Our analysis reveals that the reverse process exhibits two distinct regimes. 
During the high-to-intermediate-noise regime, the denoising model rapidly recovers the dominant forecast-relevant structure, including the overall trend and seasonality, after which the predictive statistics stabilize.
In the subsequent low-noise regime, however, successive predictions may undergo substantial statistical drift (Fig.~\ref{fig:step_predict}). 
Although the per-timestep denoising loss generally decreases with the noise level, this local improvement does not necessarily translate into a better final forecast (Fig.~\ref{fig:schedule_analysis}).
Instead, the denoising network increasingly focuses on predicting inherent noise from data, which is generally intractable.
As the inference proceeds, residual errors accumulate until the end of sampling, ultimately degrading generation quality. 
This observation exposes an important mismatch between local denoising accuracy and end-to-end forecasting performance.

\paragraph{Unequal contribution between diffusion steps.} 
We further revisit noise-schedule design. Existing time-series diffusion models often use fewer diffusion steps~\cite{d3vae,ant,d3u,cndiff} to mitigate low-noise drift, while others rely on complex architectures~\cite{diffusionts} to improve forecasting accuracy.
However, these approaches overlook the unequal contribution of different timesteps. 
Uniform timestep sampling allocates substantial model capacity to the low-noise region, despite its limited, or even harmful, role during inference.

\paragraph{Solution 1.} Motivated by these findings, we propose a simple yet effective early-stopping mechanism that terminates reverse diffusion before late-stage statistical drift emerges.
The stopping point is determined entirely from the evolution of generated-sample statistics and therefore does not require access to ground-truth future observations at inference time.
By avoiding unnecessary generation, the proposed strategy reduces sampling cost by 20--50\% while achieving comparable or better forecasting performance than the full reverse process.

\paragraph{Solution 2.} We introduce a Bernoulli timestep-sampling strategy that reallocates training emphasis toward the high-noise regime, where accurate denoising has the greatest influence on the generated forecast.
This strategy preserves the training coverage on low-noise regions while improving the model's prediction on heavily corrupted inputs.
Finally, we adopt a sufficiently long diffusion schedule together with a simple transformer-based network across all experiments. 
We show that this design reduces the sensitivity to architecture hyperparameters and noise schedules, and achieves strong forecasting performance without relying on complex architectures or elaborate guidance mechanisms.

In summary, our main contributions are as follows:
\begin{itemize}
\item We identify two distinct reverse-diffusion regimes in time series forecasting and reveal that low-noise refinement can introduce statistical drift and degrade forecast quality.

\item We propose an early-stopping mechanism that avoids harmful late-stage denoising, reducing sampling cost without requiring future ground truth.

\item We introduce a Bernoulli timestep-sampling strategy that emphasizes high-noise prediction while retaining coverage of the full diffusion trajectory.

\item Our proposed method achieves state-of-the-art probabilistic forecasting performance across diverse benchmarks.

\end{itemize}

\section{Related Work}
\subsection{Time Series Forecasting Models}

A broad range of forecasting methods has been developed to capture trend, seasonality, and multi-scale temporal structure. 
Basis-expansion models such as \texttt{N-BEATS}~\cite{nbeats} represent time series using interpretable trend and seasonal components, while \texttt{N-HiTS}~\cite{nhits} extends this idea through hierarchical interpolation at multiple resolutions. Other approaches rely on explicit decomposition or frequency-domain modeling. For example, \texttt{DLinear}~\cite{dlinear} separates seasonal and trend components before prediction, whereas \texttt{FiLM}~\cite{film} combines frequency analysis with low-rank approximation to suppress noise and preserve long-term information. 

More recently, \texttt{Transformer}-based forecasting models capture long-range dependencies through specialized architectures and tokenization strategies. \texttt{Informer}~\cite{informer}, \texttt{Autoformer}~\cite{autoformer}, and \texttt{FEDformer}~\cite{fedformer} improve efficiency and temporal modeling using sparse attention, decomposition, and frequency-domain representations. \texttt{PatchTST}~\cite{patchtst} adopts temporal patching, while \texttt{iTransformer}~\cite{itransformer} models variables as tokens. More recently, \texttt{TimesFM}~\cite{timesfm} and \texttt{Chronos}~\cite{chronos} leverage large-scale pretraining for cross-domain forecasting. However, these methods do not explicitly examine how predictive uncertainty evolves during iterative generation.

\subsection{Diffusion Models for Time Series Forecasting}

Diffusion models provide an alternative probabilistic formulation in which future trajectories are generated through iterative denoising. Existing methods improve temporal generation through conditional guidance and decomposition architectures. \texttt{TSDiff}~\cite{tsdiff} and \texttt{TimeDiT}~\cite{DBLP:journals/corr/abs-2409-02322} introduce guidance mechanisms to steer the sampling process, while \texttt{TimeDiff}~\cite{DBLP:conf/icml/ShenK23}, \texttt{CDPM}~\cite{DBLP:journals/corr/abs-2410-13253}, and \texttt{D$^3$U}~\cite{d3u} exploit temporal decomposition or structured conditioning. 
A few works like \texttt{ANT}~\cite{ant}, \texttt{CNDiff}~\cite{cndiff}, and \texttt{NsDiff}~\cite{nsdiff} consider the temporal structure of time series data in the diffusion process. 
However, the evolution of prediction quality along the reverse process remains underexplored, which motivates our further analysis of sampling dynamics.
\section{Preliminaries}
\begin{figure*}[t!]
    \centering
    \begin{subfigure}[b]{0.21\textwidth}
        \centering
        \includegraphics[width=\textwidth]{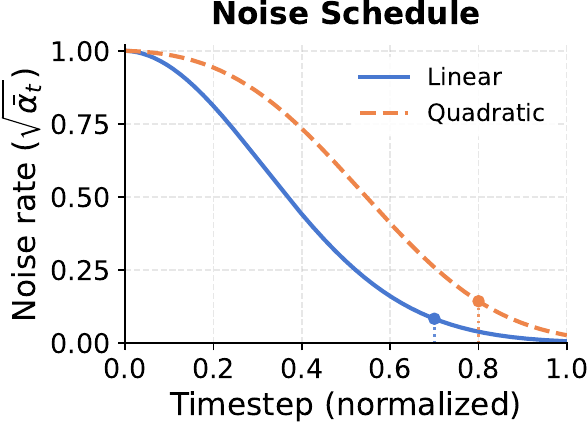}
        \caption{Noise schedule}
        \label{fig:noise_schedules}
    \end{subfigure}
    \hfill
    \begin{subfigure}[b]{0.35\textwidth}
        \centering
        \includegraphics[width=\textwidth]{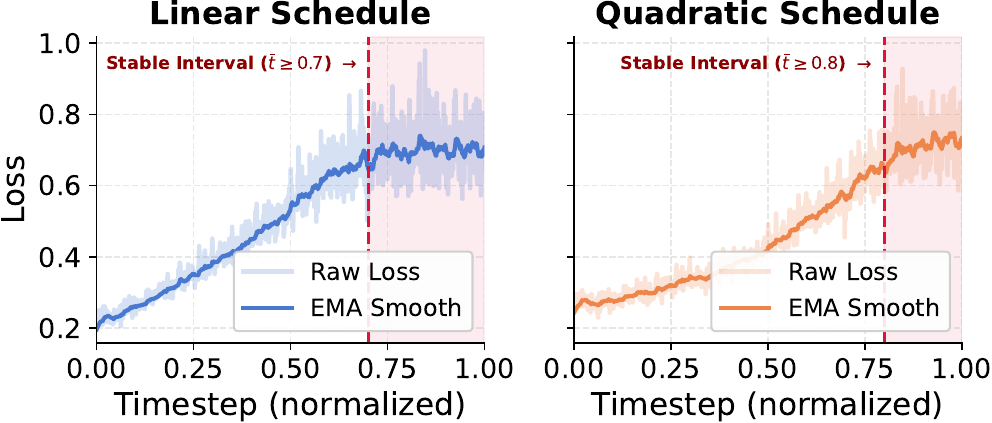}
        \caption{Per-timestep loss}
        \label{fig:timestep_losses}
    \end{subfigure}
    \hfill
    \begin{subfigure}[b]{0.40\textwidth}
        \centering
        \includegraphics[width=\textwidth]{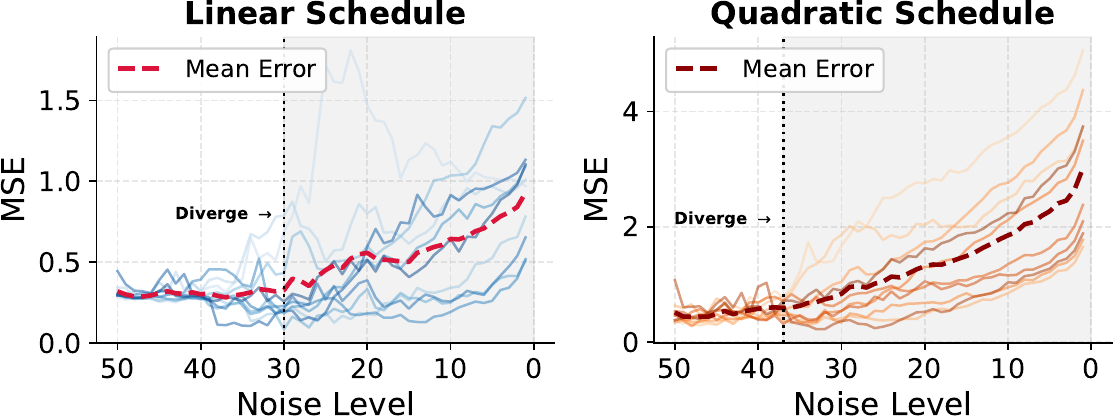}
        \caption{Candidate trajectory error}
        \label{fig:candidate_errors}
    \end{subfigure}
    \caption{\textbf{Comparison of Different Schedules.} (a) Signal scale decay $\sqrt{\bar{\alpha}_t}$ over timesteps $t$ with stability transition points. (b) Per-timestep loss curves showing convergence into the stable loss regime. (c) Candidate trajectory mean squared errors across varying noise levels.}
    \label{fig:schedule_analysis}
\end{figure*}

\paragraph{Notation.} We denote $\mathbf{x}\in\mathbb{R}^{Q\times D}$ as the historical context, $\mathbf{y}_0\in\mathbb{R}^{P\times D}$ as the future horizon, and $\mathbf{y}_t\in\mathbb{R}^{P\times D}$ as the latent at diffusion step $t\in\{0,\dots,T\}$, where $Q$, $P$, $D$ denote the context length, the prediction length, and the number of variables, respectively.
Let $\beta_t\in(0,1)$ be the noise variance schedule, $\alpha_t := 1-\beta_t$, and $\bar{\alpha}_t := \prod_{i=1}^t \alpha_i$ with $\bar{\alpha}_0=1$ and $\bar{\alpha}_T\approx0$.

\paragraph{Problem Formulation.}
The goal is to predict the future time series $\mathbf{y}_0$ given the historical context $\mathbf{x}$.
Hence, we formulate the problem as a generation task to generate $p(\mathbf{y}_0|\mathbf{x})$.
A diffusion model achieves this by iteratively running a learnable denoising process, starting from a simple Gaussian noise $p(\mathbf{y}_T)=\mathcal{N}(\mathbf{0},\mathbf{I})$:
\begin{align}
p_\theta(\mathbf{y}_{0:T}|\mathbf{x}) = p(\mathbf{y}_T)\prod_{t=1}^{T} p_\theta(\mathbf{y}_{t-1}\mid \mathbf{y}_t,\mathbf{x}).
\label{eq:reverse_chain}
\end{align}

Since $p_\theta(\mathbf{y}_{t-1}\mid \mathbf{y}_t,\mathbf{x})$ is generally intractable, we approximate it with a pre-defined posterior distribution $q(\mathbf{y}_{t-1}\mid \mathbf{y}_t,\mathbf{y}_0)$, which is derived from an auxiliary noising process:
\begin{align}
q(\mathbf{y}_t\mid \mathbf{y}_{t-1})&=\mathcal{N}\!\left(\sqrt{\alpha_t}\mathbf{y}_{t-1},\,\beta_t\mathbf{I}\right), \\ 
q(\mathbf{y}_t\mid \mathbf{y}_0)&=\mathcal{N}\!\left(\sqrt{\bar{\alpha}_t}\mathbf{y}_0,\,(1-\bar{\alpha}_t)\mathbf{I}\right),
\label{eq:forward_process} \\
q(\mathbf{y}_{t-1}\mid \mathbf{y}_t,\mathbf{y}_0)
&= \mathcal{N}\!\left(\tilde{\boldsymbol{\mu}}(\mathbf{y}_t,\mathbf{y}_0),\,\tilde{\beta}_t\mathbf{I}\right) \label{eq:posterior}
\end{align}
where $\tilde{\beta}_t := \frac{1-\bar{\alpha}_{t-1}}{1-\bar{\alpha}_t}\beta_t$ and $\tilde{\boldsymbol{\mu}}(\mathbf{y}_t,\mathbf{y}_0)
:= \frac{\sqrt{\bar{\alpha}_{t-1}}\beta_t}{1-\bar{\alpha}_t}\mathbf{y}_0
   + \frac{\sqrt{\alpha_t}(1-\bar{\alpha}_{t-1})}{1-\bar{\alpha}_t}\mathbf{y}_t$.
Since Eq.~\ref{eq:posterior} is simply a combination of $\mathbf{y}_t$ and $\mathbf{y}_0$, hence learning $p_\theta(\mathbf{y}_{t-1}\mid \mathbf{y}_t,\mathbf{x})$ reduces to approximating the clean prediction $\mathbf{y}_0$:
\begin{align}
&\mathcal{L}_{\mathrm{diffusion}}
= \mathbb{E}_{t, \mathbf{y}_0, \mathbf{y}_t}
\left[\left\|\mathbf{y}_0-\mathbf{y}_\theta(\mathbf{y}_t,\mathbf{x},t)\right\|_2^2\right],
\label{eq:simple_loss}
\end{align}
where $\mathbf{y}_t$ sampled using Eq.~\ref{eq:forward_process}.

Unlike deterministic predictors that output a single trajectory, diffusion models directly measure the \emph{full predictive distribution}. This is particularly suitable for real-world time series where (i) stochasticity and measurement noise are irreducible, and (ii) multiple plausible futures exist.
\section{Methodology}
\begin{figure*}
    \centering
    \includegraphics[width=\linewidth]{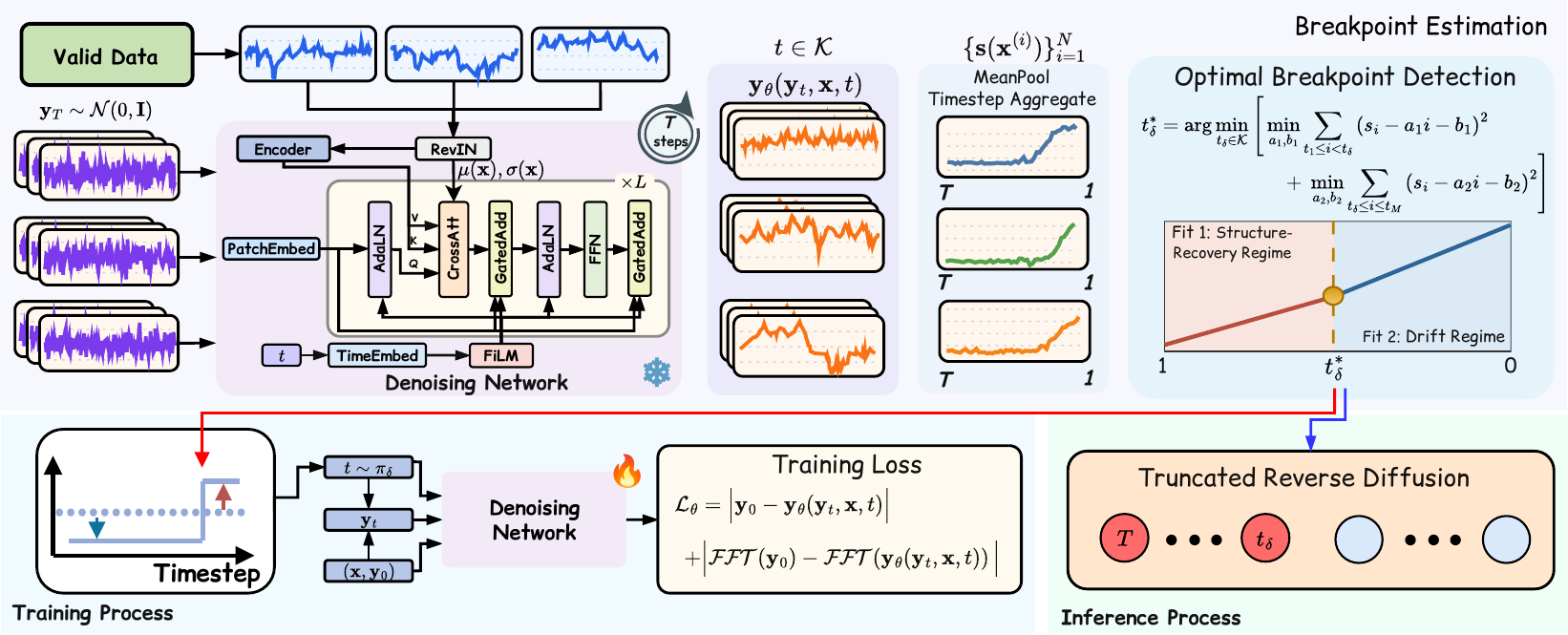}
    \caption{\textbf{Framework Overview.} 
\textbf{Breakpoint Estimation:} Prediction  trajectory of each sample $\mathbf{x}_i$ is summarized into $\mathbf{s}[\mathbf{x}_i]$, where piecewise-linear fitting detects the optimal transition breakpoint $t_\delta*$. 
\textbf{Bernoulli Sampler:} Training timesteps are adaptively sampled with probability $p$ focused on the high-noise region while retaining full trajectory coverage.
\textbf{(c) Early-Stopped Accelerated Inference:} Reverse diffusion terminates early at $t_{\delta}$, preventing error propagation and reducing compute costs.}
\label{fig:framework_overview}
\end{figure*}

Our framework aligns diffusion training and inference with the denoising dynamics of time series, as illustrated in Fig.~\ref{fig:framework_overview}. We first analyze how forecast quality evolves along the reverse trajectory and identify a transition from global structure recovery to unstable late-stage refinement (Section~\ref{sec:denoising_dynamics}). Based on this observation, we estimate a dataset-level stopping point from a small set of pilot trajectories and terminate sampling before statistical drift emerges (Section~\ref{sec:global_early_stopping}). The estimated boundary is further used to construct an adaptive Bernoulli timestep sampler that allocates more training probability to the influential high-noise region while preserving supervision over the complete diffusion horizon (Section~\ref{sec:bernoulli_sampling}).


\subsection{When Denoising Degrades Forecast Quality}
\label{sec:denoising_dynamics}
Diffusion sampling is commonly interpreted as a progressive refinement process in which every reverse step improves the generated output. Our empirical analysis shows that this assumption does not necessarily hold for time series.

Figs.~\ref{fig:noise_schedules} and~\ref{fig:timestep_losses} show the noise schedules and the denoising loss across timesteps of these schedules.
Under linear and quadratic schedules, the loss increases with the corruption level before reaching a stable high-noise interval. 
These statistics reveal that the stable interval depends on the noise schedule, where the noise is extremely high and the input signal is largely corrupted.

We further track the intermediate clean prediction
$\hat{\mathbf{y}}_0^{\,t}$ during reverse sampling.
As shown in Fig.~\ref{fig:candidate_errors}, candidate forecasts initially recover the dominant temporal structure and remain relatively close to the target. 
In the low-noise region, however, their trajectories increasingly diverge, and their errors grow toward the final step, particularly noticeable under the quadratic schedule.

These observations reveal two regimes. 
In the \emph{structure-recovery regime}, strong corruption compels the model to exploit historical context to reconstruct dominant patterns such as trend and seasonality.
In the subsequent \emph{fine-grained regime}, the intermediate sample already preserves most of this global structure, and the denoiser increasingly focuses on local residual variations.
Because these components are generally unpredictable in real-world time series and may provide limited benefit to forecasting performance~\cite{d3u}, repeatedly modeling them can introduce unnecessary variation and amplify errors across the remaining steps. We therefore refer to this late stage as the \emph{drift regime}.

This observation motivates two complementary designs: terminating generation before the generation drift and prioritizing the high-noise region during training, where denoising has the greatest influence on the resulting forecast.

\subsection{Estimating The Optimal Reverse-Diffusion Stopping Point}
\label{sec:global_early_stopping}

The generation drift varies across models, datasets, and noise schedules. 
Rather than selecting a fixed stopping timestep manually, we estimate a universal boundary from a small set of pilot trajectories and apply it to all test instances.

Although the change point can be identified using trajectory statistics such as prediction error against the ground-truth future (Fig.~\ref{fig:candidate_errors}), such criteria are unavailable at deployment.
We instead construct a ground-truth-free statistic from the expected clean prediction:
\begin{align}
    s_{t_k}(\mathbf{x}) = \mathbb{E}_{\mathbf{y}_{t_k}}\left[\mathbf{y}_{\theta}
    \left(\mathbf{y}_{t_k},\mathbf{x},t_k\right)\right],
\end{align}
where $\mathcal{K}=\{t_k\}_{k=1}^{M}$ denotes the evaluated timesteps, with
$t_1=0<\cdots<t_M=T$.
In practice, the expectation is approximated using $S$ independently sampled conditional reverse trajectories:
\begin{align}
s_{t_k}(\mathbf{x})
\approx 
\frac{1}{S}
\sum_{j=1}^{S}
\mathbf{y}_\theta
\left(
\mathbf{y}_{t_k}^{(j)},\mathbf{x},t_k
\right).\operatorname{mean()}
\label{eq:empirical_score}
\end{align}
Averaging across trajectories reduces sampling variability and prediction error.
Our sensitivity analysis in Section~\ref{subsec:ablation} further shows that the resulting boundary is insensitive to the choice of trajectory statistic.

Given the score curve $\mathbf{s}(\mathbf{x})=[s_{t_M}...s_{t_1}](\mathbf{x})$, we estimate its change point using piecewise linear regression~\cite{bai1997estimation}.
For each candidate breakpoint, we fit separate linear models to the two segments and select the breakpoint minimizing their total residual error:
\begin{align}
    t_\delta^{*}
    =
    \arg\min_{t_\delta\in\mathcal{K}}
    \Bigg[
    &\min_{a_1,b_1}
    \sum_{t_1\leq i<t_\delta}
    \left(s_i-a_1i-b_1\right)^2 \nonumber\\
    +\;&
    \min_{a_2,b_2}
    \sum_{t_M\geq i\geq t_\delta}
    \left(s_i-a_2i-b_2\right)^2
    \Bigg], \label{eq:global_breakpoint}
\end{align}
where we exclude several endpoints from either side of $\mathcal{K}$ to ensure minimum estimation steps.
We compute $t_\delta^*$ for $N$ pilot samples and use their median as the global stopping step $t_\delta$.

The boundary is estimated once before evaluation. During inference, reverse sampling terminates at $t_\delta$, and the corresponding clean prediction is returned as the forecast.
This global criterion adds negligible estimation overhead while reducing inference cost by skipping redundant reverse steps.
The breakpoint-estimation procedure is summarized in Alg.~\ref{alg:global_early_stopping}.

\subsection{Drift-Aware Bernoulli Timestep Sampling}
\label{sec:bernoulli_sampling}

Uniform timestep sampling assigns equal training probability to all noise levels, despite their unequal contributions to forecasting. 
Since our sampling process only leverages the high-noise regime while truncating the drift regime, we reallocate training toward the more influential region without completely discarding the remaining diffusion trajectory.

Let $t_{\delta}^{(e)}$ denote the drift boundary estimated from the pilot evaluation after epoch ($e-1$)-th using Eq.~\ref{eq:global_breakpoint}. We initialize $t_{\delta}^{(1)}=T$ and update it once every several epochs. 
At each training iteration, we draw $b\sim\mathrm{Bernoulli}(p)$ with probability $p$ in significant intervals and $1-p$ otherwise, and sample the timestep from
\begin{equation}
t\sim
\begin{cases}
\mathcal{U}\!\left(t_{\delta}^{(e)},T\right)
& \text{if} \ \ \ b=1,\\[2mm]
\mathcal{U}\!\left(1,t_{\delta}^{(e)}\right) & \text{if} \ \ \ b=0.
\end{cases}
\label{eq:bernoulli_timestep}
\end{equation}
Thus, $p$ controls the emphasis on highly corrupted inputs, while the complementary branch preserves coverage of the full diffusion process.
Equivalently, the induced timestep distribution is
\begin{equation}
\pi_e(t)
=
\frac{p}{T-t_{\delta}}
\mathbf{1}_{[t_{\delta},T]}(t)
+
\frac{1-p}{t_{\delta}}
\mathbf{1}_{[1,t_{\delta})}(t).
\end{equation}
Unlike hard timestep clipping~\cite{dinoiser}, this strategy retains low-noise supervision and adapts its allocation as the estimated drift boundary evolves during training.
The training process is described in Alg.~\ref{alg:bernoulli_sampling}.

\begin{algorithm}[t]
\caption{Global Breakpoint Estimation}
\label{alg:global_early_stopping}
\begin{algorithmic}[1]
\Require Denoiser $\mathbf{y}_{\theta}$, sampling schedule
$\{t_k\}_{k=1}^{M}$, number of candidates $S$, number of pilot samples $N$
\Ensure Global stopping timestep $t_{\delta}$

\State \textcolor{gray}{[Collect pilot score trajectories]}
\State $\operatorname{stop}\_\operatorname{steps}=[]$
\For{$n=1$ to $N$}
    \State $\mathbf{x}\sim q(\mathbf{x},\mathbf{y}_0)$
    \For{$j=1$ to $S$}
    \State $\mathbf{y}_{t_M}^{(j)} \sim \mathcal{N}(\mathbf{0},\mathbf{I})$
    \For{$k=M$ to $1$}
        \State Store $\mathbf{y}_{\theta}
        (\mathbf{y}_{t_k}^{(j)},\mathbf{x},t_k)$
        \If{$k>1$}
            \State $\mathbf{y}_{t_{k-1}}^{(j)}
            \sim p_{\theta}
            (\mathbf{y}_{t_{k-1}}
            \mid\mathbf{y}_{t_k}^{(j)},\mathbf{x})$
        \EndIf
    \EndFor
    \EndFor
    \State Compute $\mathbf{s}(\mathbf{x})$ following Eq.~\ref{eq:empirical_score}
    \State \textcolor{gray}{[Piecewise-linear breakpoint detection]}
    \State Estimate $t_\delta$ following Eq.~\ref{eq:global_breakpoint}
    \State $\operatorname{stop}\_\operatorname{steps}.append(t_\delta)$
\EndFor

\State $t_\delta^*=\operatorname{stop}\_\operatorname{steps}.median()$

\State \Return $t_\delta^*$
\end{algorithmic}
\end{algorithm}

\begin{algorithm}[t]
\caption{Bernoulli Distribution-Based Sampling}
\label{alg:bernoulli_sampling}
\begin{algorithmic}[1]
\Require Bernoulli parameter $p$.
\State Estimating the significant interval $t_{\delta}$ using Alg.~\ref{alg:global_early_stopping}
\Repeat
    \State $(\mathbf{x}, \mathbf{y}_0) \sim q(\mathbf{x}, \mathbf{y}_0)$
    \State Generating random numbers $u \sim \mathbb{U}[0,1]$
    \If{$u \geq p$}
        \State $t \sim \mathbb{U}(\{1,\ldots,t_{\delta}\}),\ \epsilon \sim \mathcal{N}(0,I)$
    \Else
        \State $t \sim \mathbb{U}(\{t_{\delta},\ldots,t_T\}),\ \epsilon \sim \mathcal{N}(0,I)$
    \EndIf
    \State Compute $\mathbf{y}_t$ using Eq.~\ref{eq:forward_process}
    \State  
    $\theta = \theta-\eta\nabla_{\theta}\left\|\mathbf{y}_0 - \mathbf{y}_{\theta}(\mathbf{y}_t,\mathbf{x},t)\right\|^2$
\Until{converged}
\end{algorithmic}
\end{algorithm}
\section{Experiments}
\label{sec:experiments}

\subsection{Experimental Setup}
\label{sec:experimental_setup}

\paragraph{Datasets.}
We conduct experiments on eight widely used real-world benchmarks: ETT\{h1, h2, m1, m2\}, Weather, Electricity, Exchange, and Traffic. These datasets cover diverse domains, sampling frequencies, and temporal dynamics, providing a comprehensive evaluation across different conditions. We follow the standard train--validation--test splits and pre-processing protocols used in prior works~\cite{cndiff,nsdiff}. 
The input length $Q$ is set to $192$, and results are reported over prediction lengths $P\in\{96,192,336,720\}$. Additional dataset statistics are provided in Appendix~\ref{app:datasets}.

\paragraph{Baselines.}
We evaluate deterministic and probabilistic forecasting against two complementary groups of baselines. For deterministic forecasting, we compare with representative models, including \texttt{DLinear}~\cite{dlinear}, \texttt{PatchTST}~\cite{patchtst}, \texttt{iTransformer}~\cite{itransformer}, \texttt{TimesNet}~\cite{timesnet}, \texttt{TiDE}~\cite{tide}, \texttt{TimeMixer}~\cite{timemixer}, \texttt{TimeXer}~\cite{timexer}, and \texttt{CNDiff}~\cite{cndiff}. For probabilistic forecasting, we compare with probabilistic diffusion-based forecasters, including \texttt{CSDI}~\cite{csdi}, \texttt{TMDM}~\cite{tmdm}, \texttt{D\textsuperscript{3}U}~\cite{d3u}, \texttt{CNDiff}~\cite{cndiff}, and \texttt{NsDiff}~\cite{nsdiff}. All methods are evaluated using consistent data splits, forecasting horizons, and preprocessing.

\paragraph{Metrics.}
For deterministic forecasting, we report Mean Squared Error (MSE) and Mean Absolute Error (MAE). For probabilistic forecasting, we use the Continuous Ranked Probability Score (CRPS)~\cite{crps}. 
Lower values indicate better performance for all metrics.
The detailed formula is provided in Appendix~\ref{app:datasets}.

\paragraph{Implementation details.}
We train the model using the Adam optimizer with a batch size of 64 and an initial learning rate of $0.0005$, decayed according to a cosine schedule. 
Training is performed for up to 30 epochs with early stopping.
Our model employs a Patch-based Denoising Network as described in Fig.~\ref{fig:framework_overview}, with a fixed patch size of 24.
We use $T=1000$ diffusion steps, $M=20$ sampling steps, and a \textit{linear} noise schedule, with $\beta_1=0.0001$ and $\beta_T=0.1$. 
We also utilize Fourier-transform loss from \texttt{DiffusionTS}~\cite{diffusionts} for efficient training.
For deterministic forecasting, we draw 30 samples, while for probabilistic forecasting, we draw 100 samples and set $S=30$ and $S=100$ candidate samples, correspondingly.
The number of pilot samples $N$ is set to 128. 
All experiments are conducted on a single A100 GPU with 40 GB of vRAM. 
The Bernoulli timestep-sampling parameters and adaptive stopping criterion are selected using the validation set and remain fixed during test-time evaluation.
Further architectural and optimization details are given in Appendix~\ref{app:architecture}.

\begin{table*}[t]
\centering
\small
\setlength{\tabcolsep}{2.3pt}
\renewcommand{\arraystretch}{1.15}
\begin{tabular}{l|cc|cc|cc|cc|cc|cc|cc|cc|cc}
\toprule
\textbf{Model} 
& \multicolumn{2}{c|}{\texttt{Ours}}
& \multicolumn{2}{c|}{\texttt{iTransformer}}
& \multicolumn{2}{c|}{\texttt{PatchTST}}
& \multicolumn{2}{c|}{\texttt{DLinear}}
& \multicolumn{2}{c|}{\texttt{TimesNet}}
& \multicolumn{2}{c|}{\texttt{TiDE}}
& \multicolumn{2}{c|}{\texttt{TimeMixer}}
& \multicolumn{2}{c|}{\texttt{CNDiff}}
& \multicolumn{2}{c}{\texttt{TimeXer}} \\
\midrule
\textbf{Dataset} 
& MSE & MAE
& MSE & MAE
& MSE & MAE
& MSE & MAE
& MSE & MAE
& MSE & MAE 
& MSE & MAE
& MSE & MAE 
& MSE & MAE \\
\midrule
ETTh1 
& \best{0.426} & \best{0.426} 
& 0.459 & 0.456 
& 0.499 & 0.481 
& \second{0.437} & 0.443 
& 0.534 & 0.502 
& 0.444 & 0.441 
& 0.444 & \second{0.441} 
& 0.454 & 0.461 
& 0.456 & 0.451 \\

ETTh2 
& \best{0.359} & \best{0.384} 
& 0.404 & 0.423 
& 0.380 & 0.414 
& 0.486 & 0.477 
& 0.425 & 0.439 
& 0.375 & \second{0.406} 
& \second{0.372} & 0.409 
& 0.383 & 0.420 
& 0.380 & 0.409 \\

ETTm1 
& \best{0.358} & \best{0.368} 
& 0.374 & 0.396 
& \second{0.363} & 0.390 
& 0.368 & 0.385 
& 0.531 & 0.487 
& 0.372 & \second{0.384} 
& 0.365 & 0.389 
& 0.488 & 0.452 
& 0.363 & 0.389 \\

ETTm2 
& \best{0.264} & \best{0.309} 
& 0.289 & 0.336 
& 0.277 & 0.326 
& 0.310 & 0.366 
& 0.310 & 0.346 
& 0.273 & 0.323 
& \second{0.265} & \second{0.318} 
& 0.290 & 0.344 
& 0.267 & 0.318 \\

Weather 
& \best{0.229} & \best{0.254} 
& 0.245 & 0.274 
& 0.239 & 0.268 
& 0.255 & 0.308 
& 0.282 & 0.300 
& 0.262 & 0.286 
& 0.234 & 0.271 
& 0.304 & 0.343 
& \second{0.232} & \second{0.267} \\

Electricity 
& \second{0.168} & \best{0.249} 
& 0.173 & 0.264 
& 0.177 & 0.275 
& 0.179 & 0.276 
& 0.201 & 0.303 
& 0.187 & 0.279 
& 0.169 & \second{0.259} 
& \best{0.167} & 0.265 
& 0.171 & 0.267 \\

Exchange 
& \best{0.336} & \best{0.396} 
& 0.373 & 0.418 
& 0.383 & 0.419 
& 0.373 & 0.416 
& 0.688 & 0.521 
& \second{0.363} & \second{0.406} 
& 0.396 & 0.421 
& 0.396 & 0.412 
& 0.431 & 0.432 \\

Traffic 
& \best{0.423} & \best{0.252} 
& 0.439 & 0.304 
& 0.439 & 0.297 
& 0.478 & 0.323 
& 0.629 & 0.337 
& 0.491 & 0.334 
& \second{0.436} & \second{0.296} 
& 0.587 & 0.309 
& 0.440 & \second{0.296} \\
\midrule
Average
& \best{0.320} & \best{0.330}
& 0.344 & 0.359
& 0.344 & 0.359
& 0.361 & 0.374
& 0.450 & 0.404
& 0.346 & 0.357
& \second{0.335} & \second{0.350}
& 0.384 & 0.376
& 0.342 & 0.354 \\
\bottomrule
\end{tabular}
\caption{Forecasting performance comparison. The average results of all predicted lengths are listed here. Best results are highlighted in red, and second-best results are underlined in blue.}
\label{tab:forecasting_results}
\end{table*}
\subsection{Main Results}
\label{sec:main_results}

\paragraph{Deterministic forecasting.}
Table~\ref{tab:forecasting_results} compares deterministic forecasting performance across eight benchmarks. Our method achieves the best overall results and remains highly competitive on Electricity, where its MSE is close to the SOTA \texttt{CNDiff}.

Averaged over all benchmarks, our method obtains an MSE of $0.320$ and an MAE of $0.330$, corresponding to relative improvements of $4.3\%$ and $5.8\%$ over \texttt{TimeMixer}, the strongest baseline on average. The largest gains appear on Exchange and Traffic, where MSE and MAE improve by $7.4\%$ and $13.9\%$ over the second-best results, respectively.

Compared with \texttt{CNDiff}, our method reduces average MSE and MAE by $16.4\%$ and $12.1\%$. These results suggest that recovering general temporal structure while suppressing noisy residual accumulation during denoising leads to more accurate forecasts. 
All horizon results are provided in Appendix~\ref{app:full_results}.

\begin{table}[t]
    \centering
    \small
    \setlength{\tabcolsep}{2.4pt}
    \renewcommand{\arraystretch}{1.15}

    \begin{tabular}{l|c|cccccc}
        \toprule
        \textbf{Dataset}
        & \texttt{Full}
        & \texttt{Early}
        & \texttt{CSDI}
& \texttt{TMDM}
& \texttt{D\textsuperscript{3}U}
& \texttt{NsDiff}
& \texttt{CNDiff} \\
        \midrule

        ETTh1
        & 0.303
        & \textbf{\textcolor{red}{0.306}}
        & 0.555
        & 0.385
        & \underline{\textcolor{blue}{0.314}}
        & 0.347
        & 0.421 \\

        ETTm1
        & 0.278
        & \underline{\textcolor{blue}{0.307}}
        & 0.359
        & 0.348
        & \textbf{\textcolor{red}{0.270}}
        & 0.320
        & 0.341 \\

        Weather
        & 0.150
        & \underline{\textcolor{blue}{0.161}}
        & \textbf{\textcolor{red}{0.137}}
        & 0.202
        & 0.170
        & 0.212
        & 0.219 \\

        Electricity
        & 0.176
        & \underline{\textcolor{blue}{0.209}}
        & 0.257
        & 0.463
        & \textbf{\textcolor{red}{0.174}}
        & 0.287
        & 0.213 \\

        Exchange
        & 0.178
        & \textbf{\textcolor{red}{0.183}}
        & 0.582
        & 0.228
        & 0.219
        & 0.218
        & \underline{\textcolor{blue}{0.197}} \\

        Traffic
        & 0.205
        & \underline{\textcolor{blue}{0.236}}
        & --
        & 0.557
        & \textbf{\textcolor{red}{0.199}}
        & 0.363
        & 0.300 \\

        \bottomrule
    \end{tabular}

    \caption{Probabilistic performance comparison. The results are reported with prediction length $P=96$. Due to the excessive time and memory consumption of CSDI in long-term forecasting, its results are unavailable. The best results are highlighted in \textcolor{red}{\textbf{red}}, while the second-best results are \textcolor{blue}{\underline{blue}}.    
    }
    \label{tab:crps_results}
\end{table}

\begin{table}[t]
    \centering
    \small
    \setlength{\tabcolsep}{8pt}
    \renewcommand{\arraystretch}{1.15}

    \begin{tabular}{ccccc}
        \toprule
        Early Stop
        & Bernoulli
        & MSE
        & MAE
        & CRPS  \\
        \midrule

        $\times$
        & $\times$
        & 0.450
        & 0.420
        & 0.322 \\

        $\times$
        & $\checkmark$
        & 0.412
        & 0.398
        & \textbf{0.306} \\

        $\checkmark$
        & $\times$
        & 0.339
        & 0.358
        & 0.309 \\

        $\checkmark$
        & $\checkmark$
        & \textbf{0.333}
        & \textbf{0.357}
        & 0.335 \\
        \bottomrule
    \end{tabular}

    \caption{
    Ablation study of early stopping and Bernoulli timestep sampling.
    (Bold indicates best performance).
    }
    \label{tab:early_stop_bernoulli_ablation}
\end{table}

\paragraph{Probabilistic forecasting.}
Table~\ref{tab:crps_results} compares our method with representative diffusion-based forecasting approaches in terms of CRPS. 
The results reveal a clear trade-off between forecast accuracy and sample diversity under full and early-stopped inference.
Although full reverse diffusion achieves lower CRPS than early stopping, this improvement is largely driven by increased sample dispersion caused by low-noise statistical drift, rather than by more accurate predictions.
As shown in Fig.~\ref{fig:mse_schedules}, the corresponding point forecasts remain inaccurate and unstable.
Early stopping suppresses this drift and consequently reduces excessive diversity, while preserving competitive probabilistic performance.
In particular, our method ranks first on ETTh1 and Exchange and remains among the top-performing approaches on the remaining datasets, indicating that truncating reverse diffusion achieves a favorable balance between predictive accuracy and distributional calibration.

\subsection{Ablation Studies}
\label{subsec:ablation}

\paragraph{Contribution of each component.}
Table~\ref{tab:early_stop_bernoulli_ablation} shows that both proposed components contribute to forecasting quality. Early stopping provides the largest improvement in point forecasting, confirming that late reverse steps can degrade the recovered trajectory. Bernoulli timestep sampling improves the probabilistic forecast by emphasizing the more informative high-noise region. Combining the two achieves the strongest MSE and MAE, while Bernoulli sampling alone yields the best CRPS, suggesting a mild trade-off between point accuracy and distributional calibration.

\begin{figure}[t]
    \centering

    \begin{subfigure}[t]{0.48\linewidth}
        \centering
        \includegraphics[width=\linewidth]{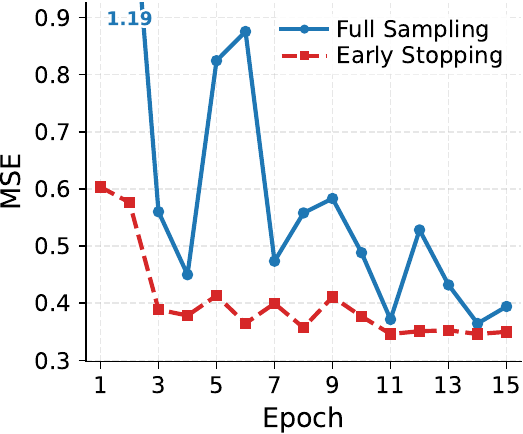}
        \caption{Linear schedule}
        \label{fig:mse_linear}
    \end{subfigure}
    \hfill
    \begin{subfigure}[t]{0.48\linewidth}
        \centering
        \includegraphics[width=\linewidth]{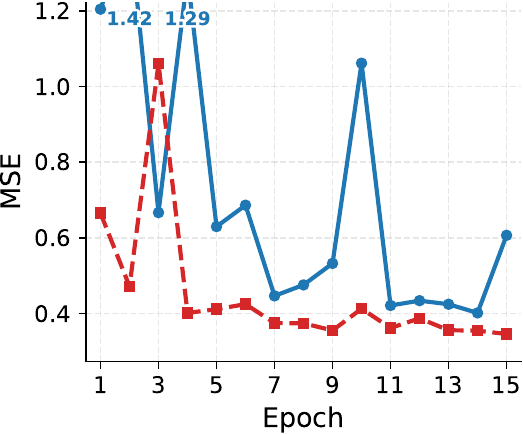}
        \caption{Quadratic schedule}
        \label{fig:mse_quadratic}
    \end{subfigure}

    \vspace{0.5em}

    \begin{subfigure}[t]{0.48\linewidth}
        \centering
        \includegraphics[width=\linewidth]{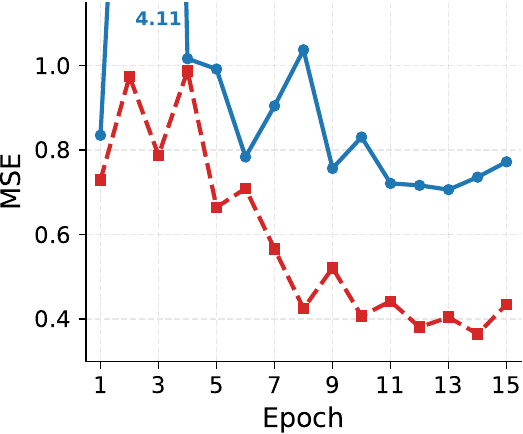}
        \caption{Cosine schedule}
        \label{fig:mse_cosine}
    \end{subfigure}
    \hfill
    \begin{subfigure}[t]{0.48\linewidth}
        \centering
        \includegraphics[width=\linewidth]{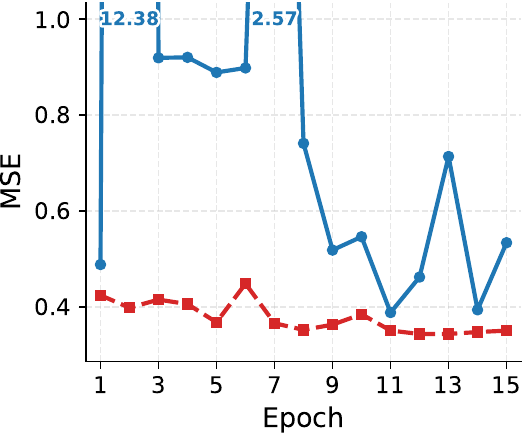}
        \caption{Sigmoid schedule}
        \label{fig:mse_sigmoid}
    \end{subfigure}

    \caption{Performance comparison across training epochs under different noise schedules.}
    \label{fig:mse_schedules}
\end{figure}

\begin{table*}[t!]
    \centering
    \small
    \setlength{\tabcolsep}{6pt}
    \renewcommand{\arraystretch}{1.18}

    \begin{tabular}{llcccccc}
        \toprule
        \textbf{Metric}
        & \textbf{Method}
        & 10
        & 20
        & 50
        & 100
        & 200
        & 1000 \\
        \midrule

        \multirow{2}{*}{MSE }
        & Full
        & 0.395
        & 0.424
        & 0.475
        & 0.516
        & 0.556
        & 0.644 \\

        & Early
        & \textbf{0.346}
        & \textbf{0.346}
        & \textbf{0.347}
        & \textbf{0.350}
        & \textbf{0.350}
        & \textbf{0.354} \\

        \midrule

        \multirow{2}{*}{Time}
        & Full
        & 1m8s
        & 2m13s
        & 4m48s
        & 9m23s
        & 17m39s
        & 86m28s \\

        & Early
        & \textbf{59s}\,
          \textcolor{savegreen}{\scriptsize (8s)}
        & \textbf{58s}\,
          \textcolor{savegreen}{\scriptsize (8s)}
        & \textbf{4m1s}\,
          \textcolor{savegreen}{\scriptsize (10s)}
        & \textbf{4m29s}\,
          \textcolor{savegreen}{\scriptsize (10s)}
        & \textbf{8m36s}\,
          \textcolor{savegreen}{\scriptsize (19s)}
        & \textbf{47m20s}\,
          \textcolor{savegreen}{\scriptsize (1m14s)} \\

        \bottomrule
    \end{tabular}%

    \caption{
    Performance comparison under different numbers of sampling steps and inference time on ETTm1 ($P=192$). 
    Green subtext reports the breakpoint estimation time.
    }
    \label{tab:ettm1_budget_comparison}
\end{table*}
\paragraph{Robustness across noise schedules.}
Fig.~\ref{fig:mse_schedules} compares full and early-stopped sampling throughout training. 
Early stopping consistently produces lower and substantially more stable MSE across all noise schedules, whereas full sampling exhibits large instability.
This suggests that poor forecasting performance does not necessarily come from noise schedule or diffusion step $T$ selection, but from the denoiser's attempts to model weakly predictable residual variations.

\paragraph{Sampling budget and efficiency.}
Table~\ref{tab:ettm1_budget_comparison} shows that increasing the sampling budget does not improve forecasting accuracy. In contrast, early-stopped sampling remains stable across all budgets. 
It reduces inference time by $13.6\%$--$56.6\%$, demonstrating that the estimated stopping point removes both harmful and computationally redundant reverse steps.
Additionally, the computational overhead of breakpoint estimation is negligible compared with the overall inference time.

\begin{table}[t]
    \centering
    \small
    \setlength{\tabcolsep}{5pt}
    \renewcommand{\arraystretch}{1.15}

    \begin{tabular}{lcccccc}
        \toprule
        \textbf{Dataset}
        & 0.7
        & 0.8
        & 0.85
        & 0.9
        & 0.95
        & 0.99 \\
        \midrule

        ETTm1
        & 0.363
        & 0.336
        & 0.344
        & \textbf{0.333}
        & 0.358
        & 0.336 \\

        Weather
        & 0.206
        & \textbf{0.203}
        & 0.205
        & 0.205
        & 0.206
        & 0.205 \\

        Exchange
        & 0.185
        & \textbf{0.171}
        & 0.175
        & \textbf{0.171}
        & 0.176
        & 0.182 \\
        \bottomrule
    \end{tabular}

    \caption{
    Sensitivity analysis (MSE) under different Bernoulli probability $p$.
    (Bold indicates best performance).
    }
    \label{tab:hyperparameter_sensitivity}
\end{table}

\paragraph{Sensitivity to Bernoulli sampling.}
As reported in Table~\ref{tab:hyperparameter_sensitivity}, performance remains stable over a broad range of $p$, with the strongest results generally obtained for $p\in[0.8,0.9]$. Smaller values allocate excessive training probability to relatively trivial low-noise timesteps, whereas overly large values provide no consistent benefit, indicating that some low-noise supervision remains necessary.

\begin{table}[t]
    \centering
    \small
    \setlength{\tabcolsep}{7pt}
    \renewcommand{\arraystretch}{1.15}

    \begin{tabular}{lcccccc}
        \toprule

        \textbf{Variance}
        & \multicolumn{2}{c}{GT Train}
        & \multicolumn{2}{c}{GT Val}
        & \multicolumn{2}{c}{Mean} \\
       
        \midrule
        \textbf{Dataset}
        & $t_{\delta}$
        & MSE
        & $t_{\delta}$
        & MSE
        & $t_{\delta}$
        & MSE  \\
        \midrule

        ETTm1
        & 350 & 0.333
        & 350 & 0.333
        & 350 & 0.333 \\

        Weather
        & 400 & 0.206
        & 350 & 0.206
        & 350 & 0.206 \\

        Exchange
        & 600 & 0.173
        & 600 & 0.173
        & 550 & 0.172 \\

        \bottomrule
    \end{tabular}

    \caption{
    Comparison of stopping points estimated using different scores.
    }
    \label{tab:trajectory_score_comparison}
\end{table}

\begin{table}[t]
    \centering
    \small
    \setlength{\tabcolsep}{5pt}
    \renewcommand{\arraystretch}{1.15}

    \begin{tabular}{lcccccc}
        \toprule

        \textbf{Dataset}
        & \multicolumn{2}{c}{ETTm1}
        & \multicolumn{2}{c}{Weather}
        & \multicolumn{2}{c}{Exchange} \\
        \midrule

        \textbf{Variance}
        & MSE 
        & MAE
        & MSE 
        & MAE 
        & MSE 
        & MAE  \\
        \midrule

        Global
        & \textbf{0.333}
        & \textbf{0.357}
        & \textbf{0.206}
        & 0.234
        & \textbf{0.172}
        & \textbf{0.295} \\

        Instance
        & 0.334
        & 0.358
        & \textbf{0.206}
        & \textbf{0.233}
        & \textbf{0.172}
        & 0.296 \\

        \bottomrule
    \end{tabular}

    \caption{
    Comparison between global and instance-level stopping-point estimation.
    (Bold indicates best performance).
    }
    \label{tab:global_vs_per_sample}
\end{table}

\paragraph{Stopping-point estimation.}
Table~\ref{tab:trajectory_score_comparison} compares the proposed label-free mean statistic with oracle criteria computed from ground-truth errors on the training or validation set. Specifically,
\begin{align}
s_{t_k}(\mathbf{x},\mathbf{y})
\approx 
\frac{1}{S}
\sum_{j=1}^{S} \left\|
\mathbf{y}_\theta
\left(
\mathbf{y}_{t_k}^{(j)},\mathbf{x},t_k
\right)-\mathbf{y}\right\|^2
\label{eq:gt_score}
\end{align}
Despite small differences in the estimated boundaries, all criteria yield nearly identical MSE. This confirms that future observations are unnecessary for reliable breakpoint estimation.
We further demonstrate the performance consistency over an interval of stopping steps in Appendix~\ref{sec:stop_sensitivity}.

\paragraph{Global versus Instance.}
We additionally conduct experiments on selecting the optimal breakpoint for individual candidates per sample.
Table~\ref{tab:global_vs_per_sample} shows that global and instance-level stopping achieve nearly identical performance across all datasets. 
We therefore adopt the global strategy, as it preserves forecasting accuracy while boosting inference time.
\section{Conclusion}

This work revisits the common assumption that every reverse-diffusion step improves time-series forecasts. Our analysis reveals that the later generation phase can induce statistical drift and degrade the final forecast. Motivated by this behavior, we introduced a label-free global stopping criterion that estimates the breakpoint from pilot trajectories and terminates sampling before harmful drift occurs. 
We further proposed a Bernoulli timestep sampler that emphasizes the more influential high-noise region while retaining supervision over the complete diffusion trajectory. Across diverse benchmarks and noise schedules, the resulting framework consistently improves point forecasting accuracy, remains competitive in probabilistic forecasting, and substantially reduces inference cost with negligible breakpoint-estimation overhead. These findings suggest that effective diffusion forecasting depends not on completing every denoising step, but on allocating computation and learning capacity to the regions that contribute most to forecast structure.


\bibliography{aaai2027}

\clearpage
\appendix

\setcounter{section}{0}
\renewcommand{\thesection}{\Alph{section}}
\setcounter{figure}{0}\renewcommand{\thefigure}{A\arabic{figure}}
\setcounter{table}{0}\renewcommand{\thetable}{A\arabic{table}}
\setcounter{equation}{0}\renewcommand{\theequation}{A\arabic{equation}}
\setcounter{algorithm}{0}\renewcommand{\thealgorithm}{A\arabic{algorithm}}
 

\newcommand{\appendixtitle}{%
  \begin{center}
    {\LARGE\bfseries
    Supplementary Material\par}
    \vspace{1.2em}
  \end{center}
}
\twocolumn[\appendixtitle]

\section{Denoising Diffusion Probabilistic Models}
\label{app:diffusion_background}

In this section, we provide a brief overview of \texttt{DDPMs} and \texttt{DDIMs}. 
A denoising diffusion probabilistic model maps the data distribution $\mathbf{y}_0\in\mathbb{R}^{P\times D}$ to a Gaussian prior $\mathbf{y}_T\in\mathbb{R}^{P\times D}$ through a fixed forward process and learns a reverse process to recover the original trajectory.

\subsection{Forward Diffusion Process}

Given a variance schedule
$\{\beta_t\}_{t=1}^{T}$, with $\beta_t\in(0,1)$, the forward diffusion process is defined as
\begin{equation}
q(\mathbf{y}_{1:T}\mid\mathbf{y}_0)
=
\prod_{t=1}^{T}
q(\mathbf{y}_t\mid\mathbf{y}_{t-1}),
\end{equation}
where $q(\mathbf{y}_t\mid\mathbf{y}_{t-1})
=
\mathcal{N}\!\left(
\sqrt{\alpha_t}\mathbf{y}_{t-1},
\beta_t\mathbf{I}
\right)$, $\alpha_t=1-\beta_t$.
Defining $\bar{\alpha}_t=\prod_{s=1}^{t}\alpha_s$, 
the noisy trajectory at any timestep can be sampled directly from $\mathbf{y}_0$:
\begin{equation}
q(\mathbf{y}_t\mid\mathbf{y}_0)
=
\mathcal{N}\!\left(
\sqrt{\bar{\alpha}_t}\mathbf{y}_0,
(1-\bar{\alpha}_t)\mathbf{I}
\right).
\label{eq:app_forward_marginal}
\end{equation}
Equivalently, $\mathbf{y}_t
=
\sqrt{\bar{\alpha}_t}\mathbf{y}_0
+
\sqrt{1-\bar{\alpha}_t}\boldsymbol{\epsilon}$, $\boldsymbol{\epsilon}\sim\mathcal{N}(\mathbf{0},\mathbf{I})$.
For a sufficiently long diffusion horizon, $\bar{\alpha}_T$ approaches zero and $\mathbf{y}_T$ becomes approximately standard Gaussian.

\subsection{Conditional Reverse Process}
For forecasting, generation is conditioned on historical context
$\mathbf{x}\in\mathbb{R}^{Q\times D}$. The reverse process is defined as
\begin{equation}
p_{\theta}(\mathbf{y}_{0:T}\mid\mathbf{x})
=
p(\mathbf{y}_T)
\prod_{t=1}^{T}
p_{\theta}
(\mathbf{y}_{t-1}\mid\mathbf{y}_t,\mathbf{x}),
\end{equation}
where
\begin{equation}
p_{\theta}
(\mathbf{y}_{t-1}\mid\mathbf{y}_t,\mathbf{x})
=
\mathcal{N}\!\left(
\boldsymbol{\mu}_{\theta}
(\mathbf{y}_t,\mathbf{x},t),
\sigma_t^2\mathbf{I}
\right).
\label{eq:app_reverse_transition}
\end{equation}

The posterior induced by the forward process is analytically tractable:
\begin{equation}
q(\mathbf{y}_{t-1}\mid\mathbf{y}_t,\mathbf{y}_0)
=
\mathcal{N}\!\left(
\tilde{\boldsymbol{\mu}}_t
(\mathbf{y}_t,\mathbf{y}_0),
\tilde{\beta}_t\mathbf{I}
\right),
\label{eq:app_forward_posterior}
\end{equation}
with $\tilde{\beta}_t
=
\frac{1-\bar{\alpha}_{t-1}}
     {1-\bar{\alpha}_t}\beta_t$ and $
\tilde{\boldsymbol{\mu}}_t
=
\frac{\sqrt{\bar{\alpha}_{t-1}}\beta_t}
     {1-\bar{\alpha}_t}\mathbf{y}_0
+
\frac{\sqrt{\alpha_t}(1-\bar{\alpha}_{t-1})}
     {1-\bar{\alpha}_t}\mathbf{y}_t$.
Thus, the reverse mean can be constructed by replacing the unknown clean $\mathbf{y}_0$ in Eq.~\ref{eq:app_forward_posterior} with a learned estimate.

We parameterize the denoiser to directly predict the clean future, using the reconstruction objective
\begin{equation*}
\mathcal{L}_{\text{diffusion}}
=
\mathbb{E}_{\mathbf{y}_0,\mathbf{x},t,\mathbf{y}_t}
\left[
w_t
\left|
\mathbf{y}_0
-
\mathbf{y}_{\theta}
(\mathbf{y}_t,\mathbf{x},t)
\right|
\right],
\label{eq:app_y0_loss}
\end{equation*}
where $\mathbf{y}_t$ is sampled from Eq.~\ref{eq:app_forward_marginal}, and $w_t$ optionally balances the contribution of different noise levels.
Unlike \texttt{DDPM}~\cite{ddpm}, which adopts an $\mathcal{L}_2$-based objective, we use an $\mathcal{L}_1$ reconstruction loss to improve robustness when modeling local temporal variations. 
We further incorporate the Fourier-transform loss introduced in \texttt{DiffusionTS}~\cite{diffusionts} to capture low-amplitude frequency components:
\begin{equation*}
\mathcal{L}_{\text{FFT}}
=
\mathbb{E}_{\mathbf{y}_0,\mathbf{x},t,\mathbf{y}_t}
\left[
\lambda_t
\left\|
\text{FFT}(\mathbf{y}_0)
-
\text{FFT}(\mathbf{y}_{\theta}
(\mathbf{y}_t,\mathbf{x},t)
\right)\|_2^2
\right],
\label{eq:app_fft_loss}
\end{equation*}
where $w_t$ and $\lambda_t$ control the contributions of the reconstruction and Fourier objectives, respectively. For simplicity, we set $w_t=\lambda_t=1$ throughout all experiments. 
The overall training objective is therefore
\begin{align}
    \mathcal{L} = \mathcal{L}_{\text{diffusion}} + \mathcal{L}_{\text{FFT}}.
\end{align}
The predicted clean trajectory is substituted into Eq.~\ref{eq:app_reverse_transition}:
\begin{equation}
\boldsymbol{\mu}_{\theta}
(\mathbf{y}_t,\mathbf{x},t)
=
\frac{\sqrt{\bar{\alpha}_{t-1}}\beta_t}
     {1-\bar{\alpha}_t}
\mathbf{y}_{\theta}
(\mathbf{y}_t,\mathbf{x},t)
+
\frac{\sqrt{\alpha_t}(1-\bar{\alpha}_{t-1})}
     {1-\bar{\alpha}_t}\mathbf{y}_t.
\label{eq:app_learned_reverse_mean}
\end{equation}

\subsection{Sampling with \texttt{DDPM}}

Generation starts from
$\mathbf{y}_T\sim\mathcal{N}(\mathbf{0},\mathbf{I})$.
At each reverse step, the model first predicts the clean trajectory and then samples
\begin{equation}
\mathbf{y}_{t-1}
=
\boldsymbol{\mu}_{\theta}
(\mathbf{y}_t,\mathbf{x},t)
+
\sigma_t\mathbf{z},
\qquad
\mathbf{z}\sim\mathcal{N}(\mathbf{0},\mathbf{I}),
\label{eq:app_reverse_sampling}
\end{equation}
where $\mathbf{z}=\mathbf{0}$ at the final step. Repeating this transition from $t=T$ to $t=1$ produces one sample from
$p_{\theta}(\mathbf{y}_0\mid\mathbf{x})$.

\subsection{Accelerated Sampling With \texttt{DDIM}}
\label{app:ddim_sampling}

To reduce the number of reverse evaluations, we additionally employ \texttt{DDIM} sampling~\cite{ddim}, which permits transitions over a subsequence of diffusion timesteps. Consider a reverse transition from timestep $t$ to an earlier timestep $s<t$. Given the clean-trajectory estimate $\hat{\mathbf{y}}_0^{\,t}
=
\mathbf{y}_{\theta}(\mathbf{y}_t,\mathbf{x},t)$, we compute the residual direction
\begin{equation*}
\mathbf{d}_t
=
\frac{
    \mathbf{y}_t
    -
    \sqrt{\bar{\alpha}_t}\hat{\mathbf{y}}_0^{\,t}
}{
    \sqrt{1-\bar{\alpha}_t}
}.
\end{equation*}
The \texttt{DDIM} transition is then given by
\begin{equation}
\mathbf{y}_s
=
\sqrt{\bar{\alpha}_s}\hat{\mathbf{y}}_0^{\,t}
+
\sqrt{1-\bar{\alpha}_s-\sigma_{t\rightarrow s}^{2}}
\,\mathbf{d}_t
+
\sigma_{t\rightarrow s}\mathbf{z},
\label{eq:app_ddim_transition}
\end{equation}
where $\mathbf{z}\sim\mathcal{N}(\mathbf{0},\mathbf{I})$ and 
$\sigma_{t\rightarrow s}
=
\eta
\sqrt{
\frac{1-\bar{\alpha}_s}
     {1-\bar{\alpha}_t}
}
\sqrt{
1-\frac{\bar{\alpha}_t}{\bar{\alpha}_s}
}$.
The parameter $\eta$ controls the stochasticity of sampling. Setting $\eta=0$ produces a deterministic trajectory, while larger values introduce additional randomness. 
In this work, we set $\eta=1$ to improve the forecasting diversity.
Because \texttt{DDIM} can skip intermediate timesteps, it enables substantially faster inference without changing the clean-trajectory prediction objective used during training.

\section{Interpretation Of Late-Stage Drift}
\label{app:theoretical_analysis}

The analysis below provides a sufficient mechanism for the observed drift; it does not imply that every diffusion forecaster necessarily degrades at low noise.

Let
$\mathbf{y}_{t-1}=\Phi_t(\mathbf{y}_t,\mathbf{x})$
denote a learned reverse transition, and let $\boldsymbol{\delta}_t$ denote the deviation from an ideal reverse trajectory.
Linearizing the transition around the ideal trajectory gives
\begin{equation}
\boldsymbol{\delta}_{t-1}
\approx
\mathbf{J}_t\boldsymbol{\delta}_t
+
\mathbf{r}_t,
\qquad
\mathbf{J}_t
=
\frac{\partial\Phi_t}{\partial\mathbf{y}_t},
\end{equation}
where $\mathbf{r}_t$ denotes the local approximation error. Hence,
\begin{equation}
|\boldsymbol{\delta}_{t-1}|
\leq
\rho_t|\boldsymbol{\delta}_t|
+
|\mathbf{r}_t|,
\qquad
\rho_t=|\mathbf{J}_t|.
\end{equation}
Unrolling the recursion from $T$ to $s$ yields
\begin{equation}
\boldsymbol{\delta}_s
\approx
\left(
\prod_{j=s+1}^{T}\mathbf{J}_j
\right)
\boldsymbol{\delta}_T
+
\sum_{k=s+1}^{T}
\left(
\prod_{j=s+1}^{k-1}\mathbf{J}_j
\right)
\mathbf{r}_k,
\end{equation}
where an empty product is defined as the identity matrix. Consequently,
\begin{equation}
|\boldsymbol{\delta}s|
\leq
\left(
\prod_{j=s+1}^{T}\rho_j
\right)
|\boldsymbol{\delta}_T|
+
\sum_{k=s+1}^{T}
\left(
\prod_{j=s+1}^{k-1}\rho_j
\right)
|\mathbf{r}_k|.
\end{equation}
When the learned and ideal reverse trajectories start from the same terminal state, 
$\boldsymbol{\delta}_T=\mathbf{0}$, the first term vanishes. When low-noise transitions become approximately identity-like, $\rho_t$ approaches one and the process ceases to be contractive. Additional reverse steps may then preserve accumulated local errors or amplify them when $\rho_t>1$. This explains why a small per-timestep denoising loss does not necessarily lead to improved end-to-end forecasting quality.

\section{Model Architecture And Training}
\label{app:architecture}

Fig.~\ref{fig:framework_overview} illustrates the model architecture.
The denoising network takes the noisy future
$\mathbf{y}_t$, historical context $\mathbf{x}$, and diffusion timestep $t$ as inputs.
We adopt a Channel-Independent design~\cite{patchtst} with \texttt{RevIN}~\cite{revin}.
The future sequence is split into non-overlapping patches of length $L_p$ and projected into $d_{\mathrm{model}}$-dimensional representation:
\begin{equation*}
\mathbf{h}^{y}_0
=
\operatorname{PatchEmbed}(\mathbf{y}_t).
\end{equation*}
The historical context is embedded separately:
\begin{equation*}
\mathbf{h}^{x}
=
\operatorname{PatchEmbed}(\mathbf{x}).
\end{equation*}

The diffusion timestep is encoded using a sinusoidal embedding followed by a multilayer perceptron. Its representation modulates each denoising block through feature-wise affine transformation:
\begin{equation*}
\operatorname{FiLM}
(\mathbf{h};t)
=
\boldsymbol{\gamma}(t)\odot\mathbf{h}
+
\boldsymbol{\beta}(t).
\end{equation*}

\begin{table}[t]
    \centering
    \setlength{\tabcolsep}{7pt}
    \renewcommand{\arraystretch}{1.15}
    \begin{tabular}{lccc}
        \hline
        \textbf{Datasets}
        & \textbf{Dropout}
        & \textbf{Encoder}
        & \textbf{Decoder} \\
        \hline
        ETTh1     & \{0.4,0.4,0.4,0.4\} & \{2,2,2,2\} & \{2,3,3,4\} \\
        ETTh2       & \{0.4,0.4,0.4,0.4\} & \{2,2,2,2\} & \{2,3,3,4\} \\
        ETTm1     & \{0.4,0.4,0.4,0.4\} & \{3,3,3,3\} & \{3,3,3,3\} \\
        ETTm2       & \{0.4,0.4,0.4,0.4\} & \{3,3,3,3\} & \{3,3,3,3\} \\
        Exchange    & \{0.2,0.2,0.2,0.2\} & \{2,2,2,2\} & \{2,2,2,2\} \\
        Weather    & \{0.4,0.4,0.4,0.2\} & \{3,3,3,3\} & \{3,3,3,4\} \\
        Electricity & \{0.2,0.2,0.2,0.2\} & \{3,3,3,3\} & \{3,3,3,3\} \\
        Traffic     & \{0.2,0.2,0.2,0.2\} & \{3,3,3,3\} & \{3,3,3,3\} \\
        \hline
    \end{tabular}
    \caption{Best hyperparameters for different datasets.
    The parameter order corresponds to prediction length \{96, 192, 336, 720\}.}
    \label{tab:model_hyperparameters}
\end{table}

The encoder and the decoder employ the Diffusion Transformer block
(\texttt{DiT}), with the diffusion timestep integrated using adaptive layer
normalization~\cite{DBLP:conf/iccv/PeeblesX23,DBLP:conf/icml/EsserKBEMSLLSBP24}. 

Each block uses cross-attention among future patches with the historical representation:
\begin{align}
\mathbf{h}^{y}_{l}
=
\operatorname{CrossAttn}
(\tilde{\mathbf{h}}^{y}_{l},\mathbf{h}^{x}).
\end{align}
The final patch representations are passed through a two-layer MLP and rearranged to produce the clean-trajectory estimate in the original forecasting space:
\begin{equation}
\mathbf{y}_{\theta}
(\mathbf{y}_t,\mathbf{x},t)
=
\operatorname{PatchDecode}
(\mathbf{h}^{y}_{L}).
\end{equation}

\paragraph{Implementation Details.}
For the model hyperparameters, we use the fixed patch length $L_p=24$, with the equal stride $S_p=24$. 
The model hidden dimension is fixed at $128$.
We use cross-attention with $4$ attention heads, followed by a feed-forward network with $ 512$-dimensional intermediate layers.
We tune only the number of historical encoder and future decoder layers, and dropout values, as we observe that higher dropout probability is beneficial for small datasets.
The configuration is shown in Tab.~\ref{tab:model_hyperparameters}.

\begin{table}[t!]
\centering
\begin{tabular}{lccccc}
\toprule
\textbf{Dataset} &
\textbf{Frequency} &
\textbf{Variables} &
\textbf{\# Observations} \\
\midrule
ETTh1       & Hourly         & 7   & 17,420 \\
ETTh2       & Hourly         & 7   & 17,420 \\
ETTm1       & 15 minutes     & 7   & 69,680 \\
ETTm2       & 15 minutes     & 7   & 69,680 \\
Weather     & 10 minutes     & 21  & 52,696 \\
Electricity & Hourly         & 321 & 26,304 \\
Exchange    & Daily          & 8   & 7,588 \\
Traffic     & Hourly         & 862 & 17,544 \\
\bottomrule
\end{tabular}
\caption{Dataset statistics and forecasting configurations.}
\label{tab:dataset_statistics}
\end{table}

For training, we use the Adam optimizer with a batch size of 64 and an initial learning rate of $0.0005$, decayed using a cosine schedule.
Models are trained for up to 30 epochs with early stopping.
We use a \textit{linear} noise schedule with $T=1000$ diffusion steps, while setting $\beta_1=0.0001$ and $\beta_T=0.1$. 
For inference, we use \texttt{DDIM} to accelerate sampling while maintaining comparable performance.
Unless otherwise specified, inference uses $M=20$ sampling steps, while other values of $M$ are evaluated throughout the paper.

\section{Implementation of Breakpoint Estimation and Bernoulli Sampling}
\label{app:stopping_details}
For breakpoint estimation, we use $N=128$ historical samples from the validation set and $S=100$ candidate samples. 
The sampling set $\mathcal{K}$ is constructed based on the number of sampling steps $M$.
In this case, we select a uniform timestep selection with $T=1000$ steps and $M=20$, hence $\mathcal{K}=\{t_{20}=1000,t_{19}=950,\ldots,t_1=50\}$.
We further exclude $10\%$ timesteps from either side of the sampling set $\mathcal{K}$, which means the total number of $t_\delta$ candidates is $16$.

\begin{figure*}[t]
    \centering

    \begin{subfigure}[t]{0.49\textwidth}
        \centering
        \begin{minipage}[t]{0.49\linewidth}
            \centering
            \includegraphics[width=\linewidth]{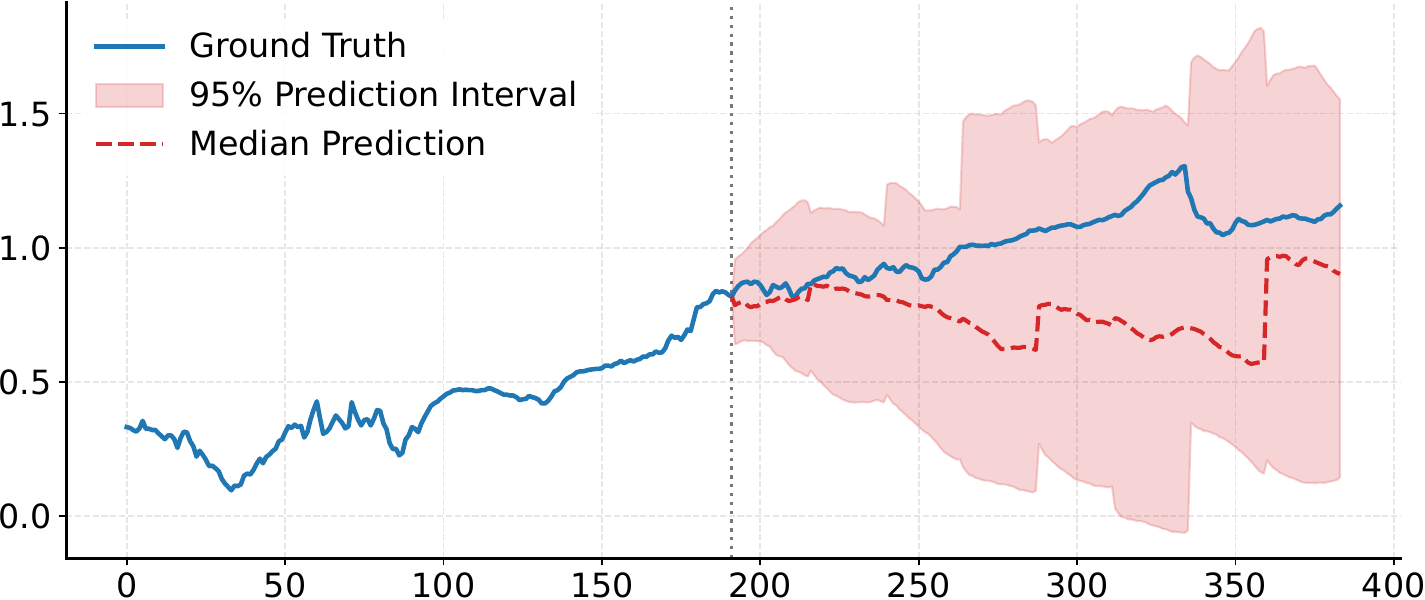}
            
            \caption*{\small Full}
        \end{minipage}
        \hfill
        \begin{minipage}[t]{0.49\linewidth}
            \centering
            \includegraphics[width=\linewidth]{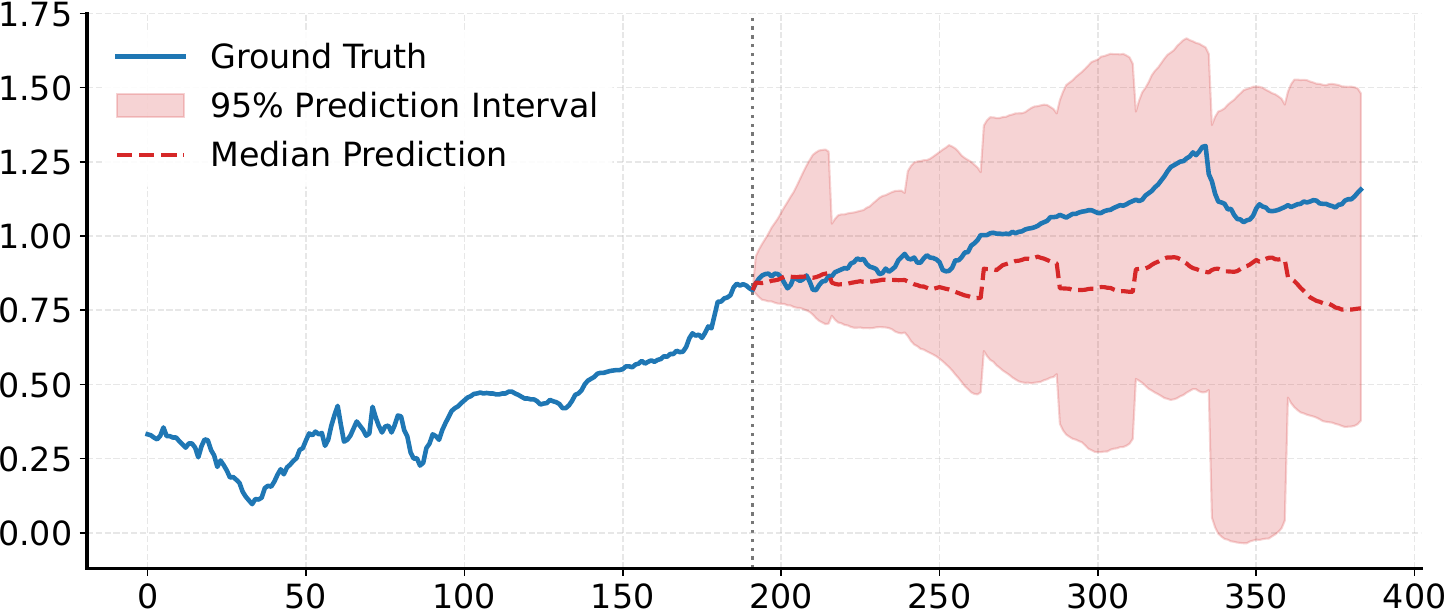}
            
            \caption*{\small Truncated}
        \end{minipage}
        
        \caption{Weather}
        \label{fig:weather_comparison}
    \end{subfigure}
    \hfill
    \begin{subfigure}[t]{0.49\textwidth}
        \centering
        \begin{minipage}[t]{0.49\linewidth}
            \centering
            \includegraphics[width=\linewidth]{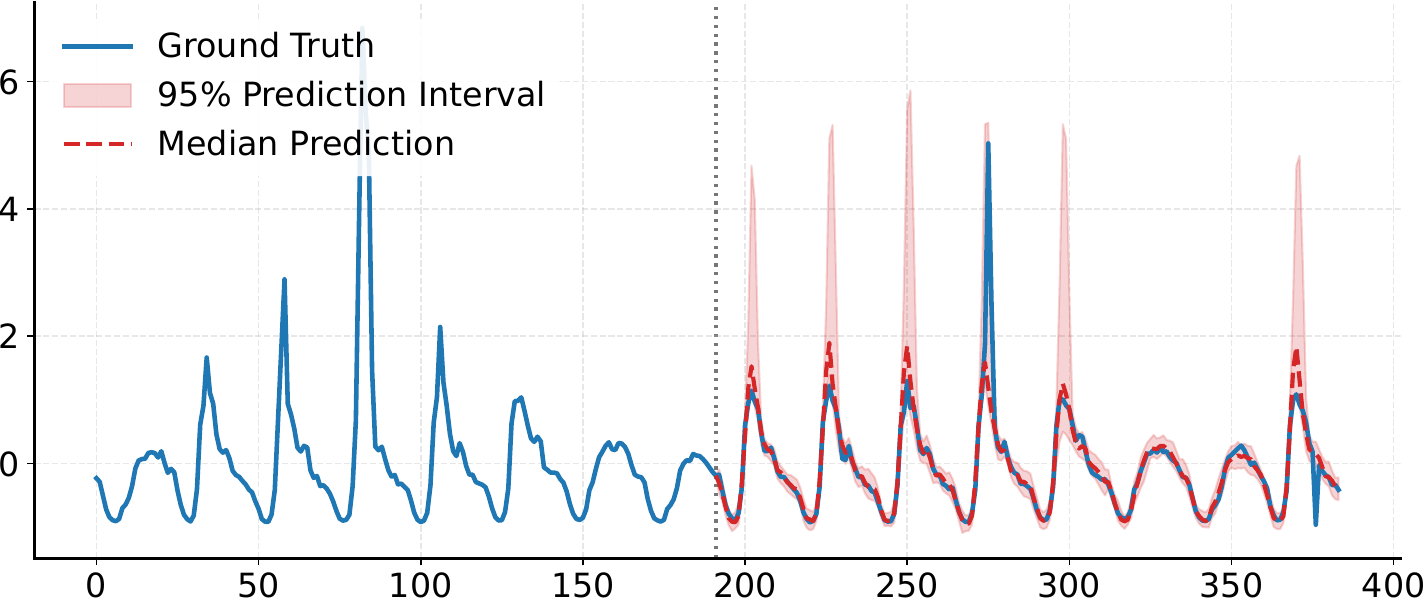}
            \caption*{\small Full}
        \end{minipage}
        \hfill
        \begin{minipage}[t]{0.49\linewidth}
            \centering
            \includegraphics[width=\linewidth]{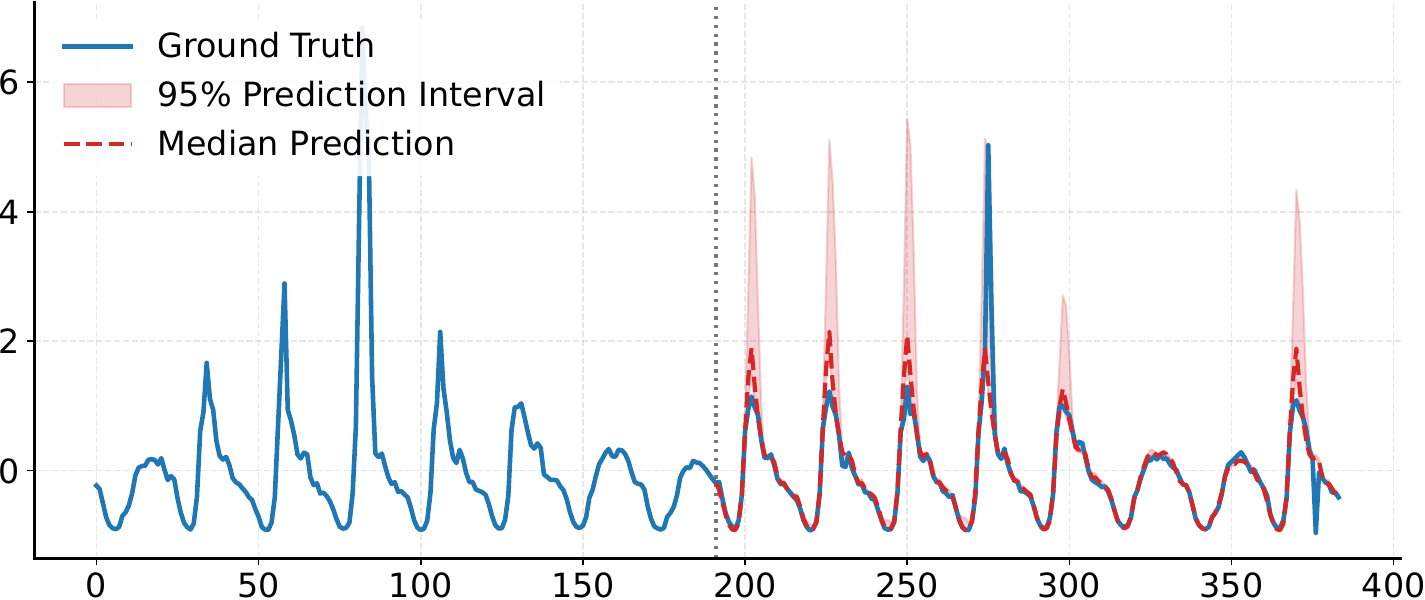}
            \caption*{\small Truncated}
        \end{minipage}
        \caption{Traffic}
        \label{fig:traffic_comparison}
    \end{subfigure}

    \begin{subfigure}[t]{0.49\textwidth}
        \centering
        \begin{minipage}[t]{0.49\linewidth}
            \centering
            \includegraphics[width=\linewidth]{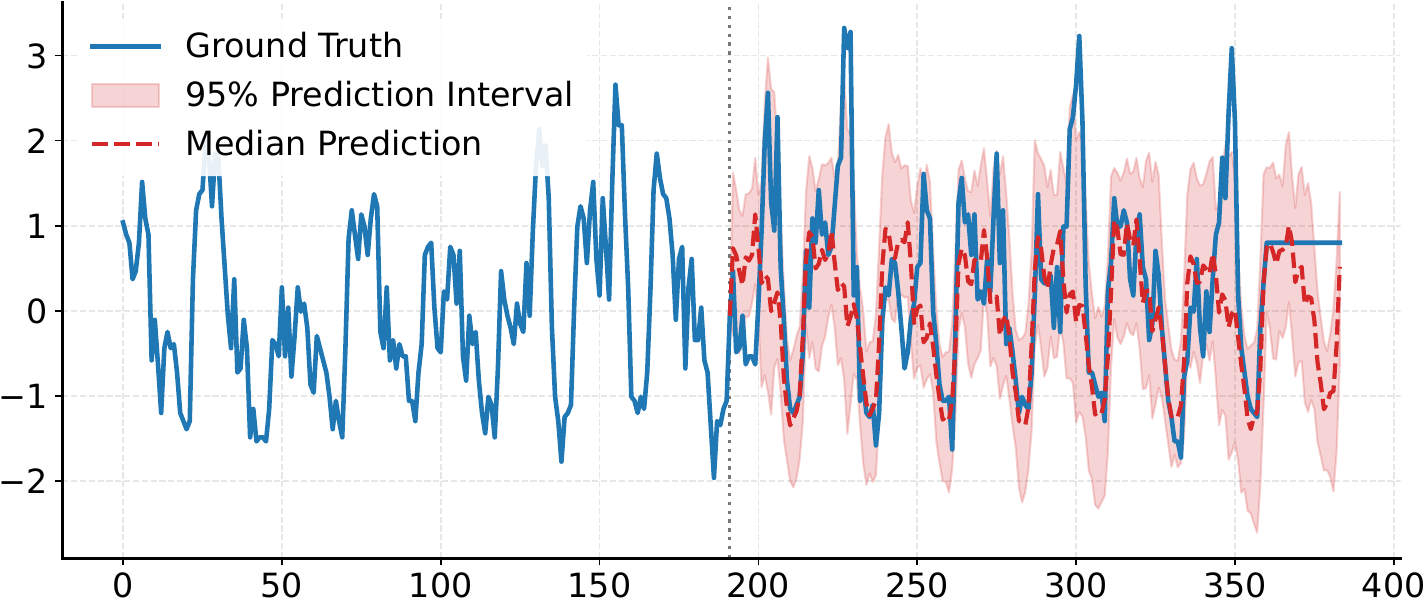}
            
            \caption*{\small Full}
        \end{minipage}
        \hfill
        \begin{minipage}[t]{0.49\linewidth}
            \centering
            \includegraphics[width=\linewidth]{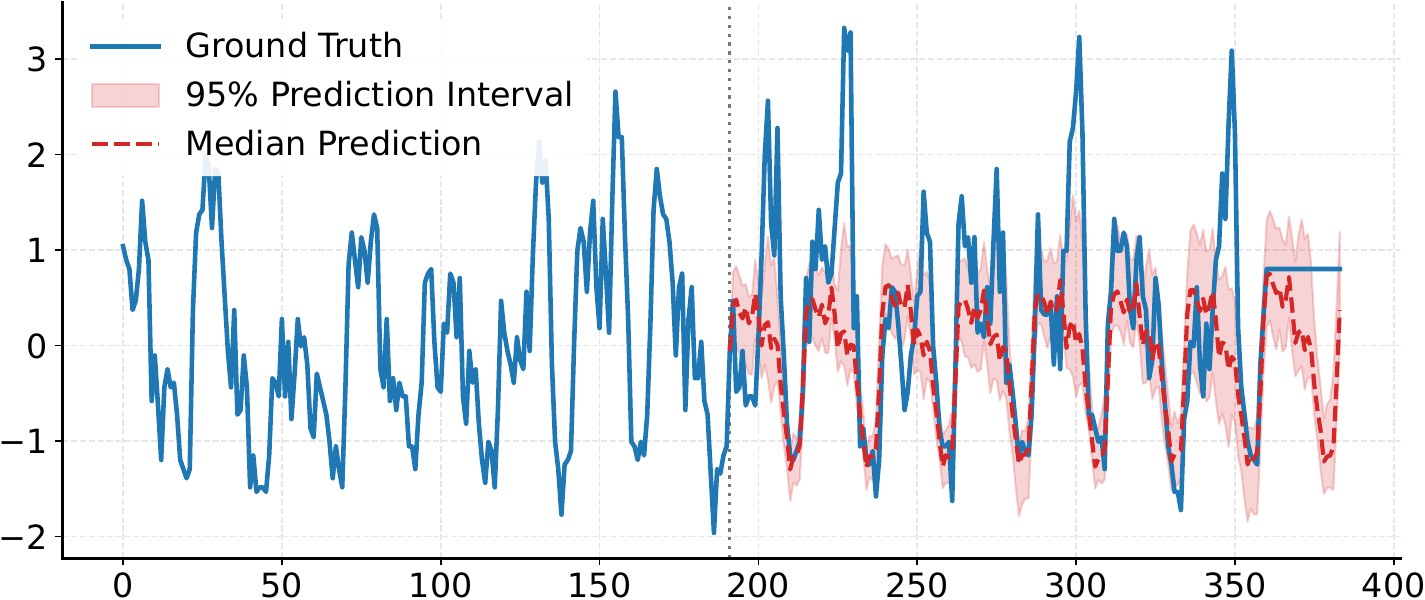}
            
            \caption*{\small Truncated}
        \end{minipage}
        
        \caption{ETTh1}
        \label{fig:etth1_comparison}
    \end{subfigure}
    \hfill
    \begin{subfigure}[t]{0.49\textwidth}
        \centering
        \begin{minipage}[t]{0.49\linewidth}
            \centering
            \includegraphics[width=\linewidth]{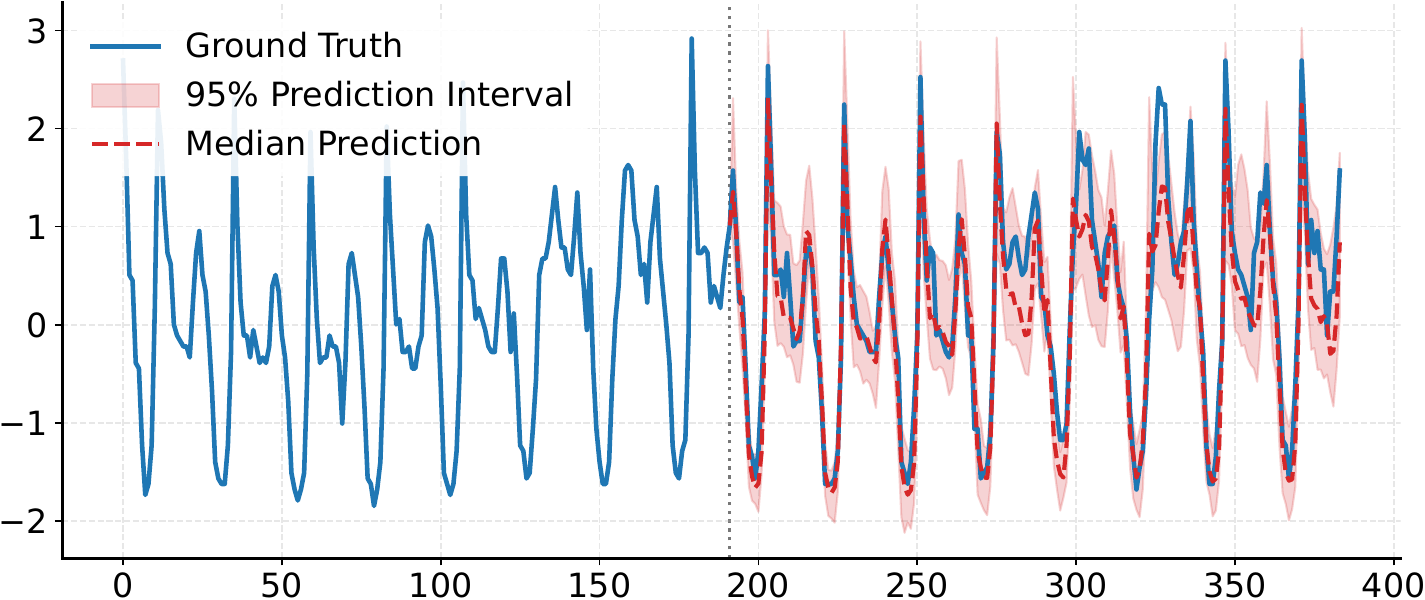}
            
            \caption*{\small Full}
        \end{minipage}
        \hfill
        \begin{minipage}[t]{0.49\linewidth}
            \centering
            \includegraphics[width=\linewidth]{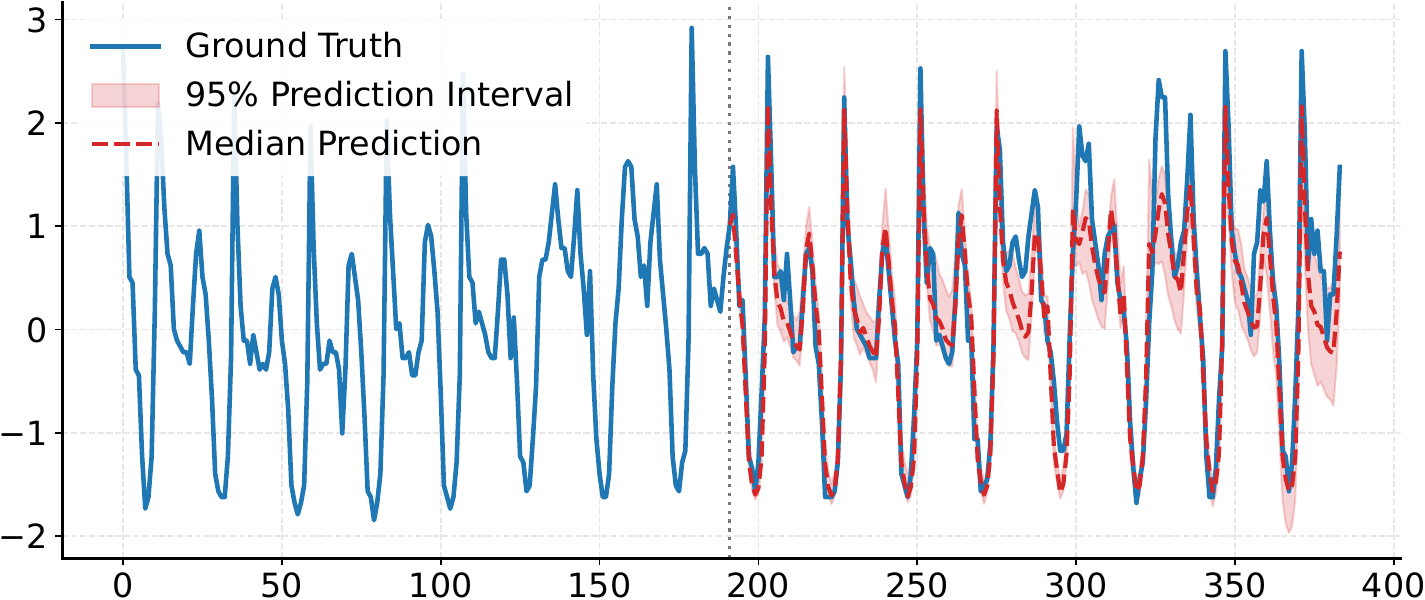}
            
            \caption*{\small Truncated}
        \end{minipage}
        
        \caption{Electricity}
        \label{fig:electricity_comparison}
    \end{subfigure}

    \caption{Qualitative comparison of probabilistic forecasts obtained using
    the full and truncated reverse diffusion processes. The shaded regions
    represent the 95\% prediction intervals.}
    \label{fig:full_trunc_visualization}
\end{figure*}

For the Bernoulli sampler, we set the mixing probability to $p=0.9$, assigning greater sampling probability to the high-noise region. After each training epoch, the boundary $t_\delta$ is updated using the proposed breakpoint-estimation algorithm. 
As shown in Tab.~\ref{tab:ettm1_budget_comparison}, this update introduces negligible computational overhead compared with the overall training and inference costs.

\section{Datasets And Evaluation Protocol}
\label{app:datasets}

We evaluate on ETTh1, ETTh2, ETTm1, ETTm2, Weather, Electricity, Exchange, and Traffic. 
The statistics for each dataset are demonstrated in Tab.~\ref{tab:dataset_statistics}.
The ETT datasets are split 12/4/4 months for train/val/test, while others are split 7:1:2. 
Statistics are computed using the training partition only, and each variable is standardized to zero mean and unit variance. 
Before evaluation, predictions are standardized given the training data statistics, following previous works.

\paragraph{Deterministic forecasts.}
Given $S$ generated trajectories, we use their sample mean as the point prediction:
\begin{equation}
\bar{\mathbf{y}}
=
\frac{1}{S}\sum_{s=1}^{S}\hat{\mathbf{y}}^{(s)}.
\end{equation}
We report
\begin{align}
\operatorname{MSE}
&=
\frac{1}{PD}
\left\|
\bar{\mathbf{y}}-\mathbf{y}_0
\right\|_2^2,\\
\operatorname{MAE}
&=
\frac{1}{PD}
\left|
\bar{\mathbf{y}}-\mathbf{y}_0
\right|.
\end{align}

\paragraph{Probabilistic forecasts.}
For predictive samples
$\{\hat{\mathbf{y}}^{(s)}\}_{s=1}^{S}$ and observation $\mathbf{y}$, we compute empirical CRPS as
\begin{equation}
\operatorname{CRPS}
=
\frac{1}{S}
\sum_{s=1}^{S}
\left|
\hat{\mathbf{y}}^{(s)}-\mathbf{y}
\right|
-
\frac{1}{2S^2}
\sum_{s=1}^{S}
\sum_{s'=1}^{S}
\left|
\hat{\mathbf{y}}^{(s)}-\hat{\mathbf{y}}^{(s')}
\right|.
\label{eq:empirical_crps}
\end{equation}
The final score is averaged over all samples, prediction steps, and variables.

\section{Additional Ablations And Analyses}
\label{app:additional_experiments}

In this section, we present additional experiments that we omitted in the main body of the paper due to limited space.
\subsection{Qualitative forecasts}

Fig.~\ref{fig:full_trunc_visualization} compares forecasts obtained using the complete and truncated reverse processes on four representative datasets. The full sampler often produces increasingly dispersed predictive intervals and median trajectories that deviate from the dominant temporal pattern, particularly on Weather and ETTh1. In contrast, early-stopped sampling yields more concentrated uncertainty estimates and better preserves the trend, periodicity, and peak structure of the ground-truth sequence. Similar behavior is observed on Traffic and Electricity, where truncation maintains the recurring temporal patterns while reducing unnecessary variation around the median prediction.

\subsection{Noise schedule design}

Tab.~\ref{tab:schedule_comparison} evaluates full and early-stopped sampling under different diffusion schedules. 
We mainly adjust the noise schedule and the diffusion steps $T$. 
Our method consistently improves both MSE and MAE across all datasets and configurations, demonstrating that its effectiveness is not tied to a particular schedule or diffusion steps. 
In contrast, full sampling does not benefit consistently from increasing $T$ and can substantially degrade performance, especially on Exchange, indicating that additional low-noise refinement may amplify forecasting errors rather than correct them. 
The relatively stable performance of our method across different values of $T$ and noise schedule further gives more options for choosing an appropriate design, without being tied to a narrow design as in previous works.

\subsection{Sensitivity to stopping-point estimation}
\label{sec:stop_sensitivity}

\begin{figure}
    \centering
    \includegraphics[width=\linewidth]{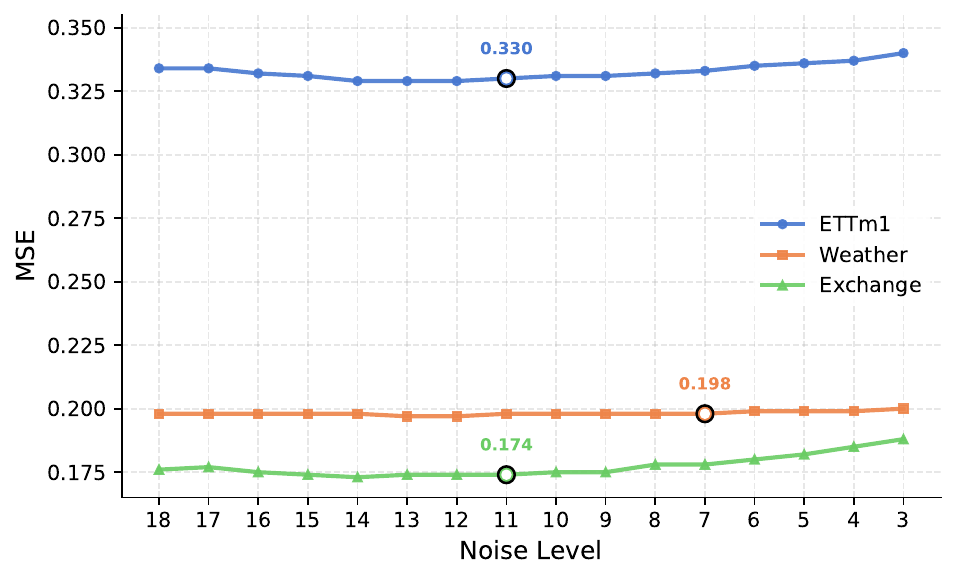}
    \caption{Sensitivity analysis on ETTm1, Weather, and Exchange datasets with stopping steps highlighted. The noise level is scaled down by a factor of $T/M=50$. The estimated stopping point is indicated by a circle.}
    \label{fig:sensitivity_anal}
\end{figure}

\begin{table*}[t]
    \centering
    \small
    \setlength{\tabcolsep}{10pt}
    \renewcommand{\arraystretch}{1.1}
    \begin{tabular}{lccccccc}
        \toprule
        \textbf{Dataset}
        & Save (\%)
        & \texttt{Early}
        & \texttt{CSDI}
        & \texttt{TMDM}
        & \texttt{D$^3$U}
        & \texttt{NsDiff}
        & \texttt{CNDiff} \\
        \midrule
        ETTh1
        & 55
        & 0.109
        & 4.874
        & 1.141
        & 0.172
        & 0.067
        & 0.425 \\

        ETTm1
        & 55
        & 0.106
        & 4.892
        & 1.141
        & 0.170
        & 0.081
        & 0.313 \\

        Weather
        & 35
        & 0.085
        & 15.412
        & 1.125
        & 0.251
        & 0.078
        & 0.263 \\

        Exchange
        & 55
        & 0.079
        & 6.020
        & 1.137
        & 0.170
        & 0.070
        & 0.275 \\

        Electricity
        & 45
        & 0.916
        & 199.882
        & 2.905
        & 1.770
        & 0.122
        & 1.864 \\

        Traffic
        & 40
        & 2.280
        & 635.500
        & 5.266
        & 2.507
        & 1.238
        & 4.890 \\
        \bottomrule
    \end{tabular}

    \caption{
Inference time (s) for generating 1 sample with prediction length $P=192$, using $S=100$ candidates.
\emph{Save} denotes the percentage of reverse-sampling computation removed by early stopping relative to the complete trajectory.
}
    \label{tab:inference_time}
\end{table*}
Fig.~\ref{fig:sensitivity_anal} evaluates forecasting performance when sampling is terminated at different noise levels. The MSE curves remain relatively flat within a broad neighborhood around the estimated stopping point, indicating that the method is not sensitive to small estimation errors. The selected boundary achieves near-optimal performance on all three datasets, with MSE values of $0.330$, $0.198$, and $0.174$ on ETTm1, Weather, and Exchange, respectively. Performance gradually deteriorates when sampling proceeds substantially beyond this region, particularly on Exchange, supporting our observation that further denoising at low-noise level can introduce forecast drift. These results suggest that the proposed estimator identifies a stable termination region on the three evaluated datasets.

\subsection{Inference Time}
Tab.~\ref{tab:inference_time} reports the inference time required to generate one forecast sample with prediction length $P=192$, using $S=100$ candidates. 
By early stopping, our method removes approximately $35$--$55\%$ of the reverse-sampling computation across datasets. It consistently achieves lower inference time than \texttt{CSDI}, \texttt{TMDM}, \texttt{D}$^3$\texttt{U}, and \texttt{CNDiff}, with particularly substantial advantages on the high-dimensional Electricity and Traffic datasets. Although \texttt{NsDiff} is slightly faster in absolute runtime, our method remains highly efficient while providing stronger overall forecasting performance. 
The later stopping points on Weather, Electricity, and Traffic indicate that these datasets benefit from more reverse steps under the current model and noise schedule.

\begin{table*}[t]
\centering
\setlength{\tabcolsep}{7pt}
\renewcommand{\arraystretch}{1.15}
\begin{tabular}{lllcccccc}
\toprule
\multicolumn{3}{l}{} 
& \multicolumn{2}{c}{ETTm1}
& \multicolumn{2}{c}{Weather}
& \multicolumn{2}{c}{Exchange} \\
\cmidrule(lr){4-5}
\cmidrule(lr){6-7}
\cmidrule(lr){8-9}
Schedule & Step & Metric
& Full & Ours
& Full & Ours
& Full & Ours \\
\midrule

\multirow{8}{*}{Linear}
& $T=20$      & MSE & 0.353 & \textbf{0.331} & 0.201 & \textbf{0.200}   & 0.274 & \textbf{0.183} \\
& $\beta_T=0.5$  & MAE & 0.373 & \textbf{0.356} & 0.240  & \textbf{0.234} & 0.394 & \textbf{0.314} \\
\cmidrule(lr){2-9}
& $T=50$      & MSE & 0.350  & \textbf{0.333} & 0.214 & \textbf{0.200}   & 0.178 & \textbf{0.168} \\
& $\beta_T=0.3$  & MAE & 0.370  & \textbf{0.358} & 0.249 & \textbf{0.235} & 0.302 & \textbf{0.295} \\
\cmidrule(lr){2-9}
& $T=100$     & MSE & 0.363 & \textbf{0.327} & 0.201 & \textbf{0.198} & 0.200   & \textbf{0.167} \\
& $\beta_T=0.1$  & MAE & 0.380  & \textbf{0.355} & 0.239 & \textbf{0.235} & 0.328 & \textbf{0.294} \\
\cmidrule(lr){2-9}
& $T=100$0    & MSE & 0.345 & \textbf{0.331} & 0.203 & \textbf{0.198} & 0.199 & \textbf{0.175} \\
& $\beta_T=0.02$ & MAE & 0.368 & \textbf{0.355} & 0.245 & \textbf{0.233} & 0.320  & \textbf{0.300}   \\
\midrule

\multirow{8}{*}{Quadratic}
& $T=20$      & MSE & 0.354 & \textbf{0.332} & 0.204 & \textbf{0.198} & 0.247 & \textbf{0.172} \\
& $\beta_T=0.6$  & MAE & 0.376 & \textbf{0.360}  & 0.252 & \textbf{0.235} & 0.374 & \textbf{0.303} \\
\cmidrule(lr){2-9}
& $T=50$      & MSE & 0.357 & \textbf{0.329} & 0.206 & \textbf{0.199} & 0.198 & \textbf{0.168} \\
& $\beta_T=0.5$  & MAE & 0.380  & \textbf{0.353} & 0.249 & \textbf{0.236} & 0.334 & \textbf{0.300}   \\
\cmidrule(lr){2-9}
& $T=100$     & MSE & 0.349 & \textbf{0.33}  & 0.206 & \textbf{0.198} & 0.258 & \textbf{0.174} \\
& $\beta_T=0.2$  & MAE & 0.366 & \textbf{0.356} & 0.246 & \textbf{0.233} & 0.375 & \textbf{0.297} \\
\cmidrule(lr){2-9}
& $T=100$0    & MSE & 0.363 & \textbf{0.334} & 0.205 & \textbf{0.200}   & 0.284 & \textbf{0.172} \\
& $\beta_T=0.02$ & MAE & 0.385 & \textbf{0.359} & 0.238 & \textbf{0.233} & 0.397 & \textbf{0.303} \\
\midrule

\multirow{8}{*}{Cosine}
& $T=20$   & MSE & 0.349 & \textbf{0.332} & 0.202 & \textbf{0.199} & 0.210  & \textbf{0.178} \\
&        & MAE & 0.365 & \textbf{0.355} & 0.240  & \textbf{0.235} & 0.352 & \textbf{0.306} \\
\cmidrule(lr){2-9}
& $T=50$   & MSE & 0.361 & \textbf{0.334} & 0.200   & \textbf{0.198} & 0.203 & \textbf{0.175} \\
&        & MAE & 0.377 & \textbf{0.357} & 0.237 & \textbf{0.232} & 0.342 & \textbf{0.307} \\
\cmidrule(lr){2-9}
& $T=100$  & MSE & 0.360  & \textbf{0.335} & 0.208 & \textbf{0.201} & 0.233 & \textbf{0.176} \\
&        & MAE & 0.377 & \textbf{0.357} & 0.256 & \textbf{0.238} & 0.361 & \textbf{0.304} \\
\cmidrule(lr){2-9}
& $T=100$0 & MSE & 0.348 & \textbf{0.332} & 0.201 & \textbf{0.198} & 0.258 & \textbf{0.174} \\
&        & MAE & 0.363 & \textbf{0.355} & 0.239 & \textbf{0.232} & 0.375 & \textbf{0.304} \\
\bottomrule
\end{tabular}
\caption{Forecasting performance under different noise schedules and diffusion steps. The prediction length is set to $P=192$. Bold values indicate better performance.}
\label{tab:schedule_comparison}
\end{table*}

\section{Full Results}
\label{app:full_results}
Tabs.~\ref{tab:ett_full_results} and~\ref{tab:full_remain} report the performance on every forecasting length setting.
\begin{table*}[t]
    \centering
    \scriptsize
    \setlength{\tabcolsep}{2.2pt}
    \renewcommand{\arraystretch}{1.08}

    \begin{tabular}{cc c|c|c|c|c|c|c|c|c|c}
        \toprule
        \textbf{Dataset}
        & \textbf{Metric}
        & \textbf{$P$}
        & \texttt{Ours}
& \texttt{iTransformer}
& \texttt{PatchTST}
& \texttt{DLinear}
& \texttt{TimesNet}
& \texttt{TiDE}
& \texttt{TimeMixer}
& \texttt{CNDiff}
& \texttt{TimeXer} \\
        \midrule

        \multirow{10}{*}{ETTh1}
        & \multirow{5}{*}{MSE}
        & 96
        & \best{0.365}
        & 0.393
        & 0.415
        & \second{0.381}
        & 0.460
        & 0.390
        & 0.389
        & 0.416
        & 0.390 \\

        & & 192
        & \best{0.418}
        & 0.441
        & 0.464
        & \second{0.424}
        & 0.480
        & 0.434
        & 0.444
        & 0.450
        & 0.430 \\

        & & 336
        & \best{0.452}
        & 0.486
        & 0.522
        & \second{0.457}
        & 0.532
        & 0.477
        & 0.467
        & 0.472
        & 0.475 \\

        & & 720
        & \best{0.468}
        & 0.516
        & 0.593
        & 0.484
        & 0.663
        & \second{0.473}
        & 0.475
        & 0.477
        & 0.530 \\

        \cmidrule(lr){3-12}
        & & avg
        & \best{0.426}
        & 0.459
        & 0.499
        & \second{0.437}
        & 0.534
        & 0.444
        & 0.444
        & 0.454
        & 0.456 \\

        \cmidrule(lr){2-12}

        & \multirow{5}{*}{MAE}
        & 96
        & \best{0.392}
        & 0.412
        & 0.428
        & \second{0.401}
        & 0.453
        & 0.404
        & 0.406
        & 0.430
        & 0.406 \\

        & & 192
        & \best{0.422}
        & 0.440
        & 0.462
        & \second{0.425}
        & 0.469
        & 0.428
        & 0.434
        & 0.452
        & 0.429 \\

        & & 336
        & \best{0.433}
        & 0.466
        & 0.491
        & \second{0.447}
        & 0.501
        & 0.455
        & 0.450
        & 0.472
        & 0.462 \\

        & & 720
        & \best{0.457}
        & 0.506
        & 0.543
        & 0.497
        & 0.585
        & 0.476
        & \second{0.472}
        & 0.491
        & 0.508 \\

        \cmidrule(lr){3-12}
        & & avg
        & \best{0.426}
        & 0.456
        & 0.481
        & 0.443
        & 0.502
        & 0.441
        & \second{0.441}
        & 0.461
        & 0.451 \\

        \midrule

        \multirow{10}{*}{ETTh2}
        & \multirow{5}{*}{MSE}
        & 96
        & \best{0.286}
        & 0.314
        & \second{0.297}
        & 0.317
        & 0.407
        & 0.304
        & 0.302
        & 0.301
        & 0.309 \\

        & & 192
        & \best{0.358}
        & 0.399
        & 0.372
        & 0.427
        & 0.425
        & 0.380
        & \second{0.359}
        & 0.364
        & 0.374 \\

        & & 336
        & \best{0.386}
        & 0.427
        & 0.426
        & 0.482
        & 0.423
        & \second{0.395}
        & 0.401
        & 0.406
        & 0.413 \\

        & & 720
        & \best{0.405}
        & 0.475
        & 0.426
        & 0.719
        & 0.445
        & \second{0.420}
        & 0.426
        & 0.460
        & 0.423 \\

        \cmidrule(lr){3-12}
        & & avg
        & \best{0.359}
        & 0.404
        & 0.380
        & 0.486
        & 0.425
        & 0.375
        & \second{0.372}
        & 0.383
        & 0.380 \\

        \cmidrule(lr){2-12}

        & \multirow{5}{*}{MAE}
        & 96
        & \best{0.330}
        & 0.362
        & \second{0.356}
        & 0.376
        & 0.417
        & \second{0.356}
        & 0.357
        & 0.352
        & 0.357 \\

        & & 192
        & \best{0.378}
        & 0.416
        & 0.407
        & 0.448
        & 0.435
        & 0.404
        & \second{0.395}
        & 0.398
        & 0.404 \\

        & & 336
        & \best{0.402}
        & 0.438
        & 0.441
        & 0.481
        & 0.440
        & \second{0.421}
        & 0.422
        & 0.440
        & 0.429 \\

        & & 720
        & \best{0.425}
        & 0.474
        & 0.450
        & 0.604
        & 0.462
        & \second{0.444}
        & 0.462
        & 0.489
        & 0.446 \\

        \cmidrule(lr){3-12}
        & & avg
        & \best{0.384}
        & 0.423
        & 0.414
        & 0.477
        & 0.439
        & \second{0.406}
        & 0.409
        & 0.420
        & 0.409 \\

        \midrule

        \multirow{10}{*}{ETTm1}
        & \multirow{5}{*}{MSE}
        & 96
        & \best{0.289}
        & 0.302
        & \second{0.299}
        & 0.310
        & 0.414
        & 0.316
        & 0.307
        & 0.328
        & 0.305 \\

        & & 192
        & \second{0.333}
        & 0.349
        & 0.338
        & 0.344
        & 0.497
        & 0.347
        & \best{0.332}
        & 0.754
        & 0.344 \\

        & & 336
        & \second{0.381}
        & 0.391
        & 0.381
        & 0.382
        & 0.612
        & 0.383
        & \best{0.378}
        & 0.439
        & 0.382 \\

        & & 720
        & \second{0.428}
        & 0.454
        & 0.432
        & 0.434
        & 0.601
        & 0.443
        & 0.443
        & 0.432
        & \best{0.422} \\

        \cmidrule(lr){3-12}
        & & avg
        & \best{0.358}
        & 0.374
        & \second{0.363}
        & 0.368
        & 0.531
        & 0.372
        & 0.365
        & 0.488
        & 0.363 \\

        \cmidrule(lr){2-12}

        & \multirow{5}{*}{MAE}
        & 96
        & \best{0.324}
        & 0.353
        & \second{0.349}
        & \second{0.349}
        & 0.422
        & 0.352
        & 0.351
        & 0.372
        & \second{0.349} \\

        & & 192
        & \best{0.351}
        & 0.381
        & 0.373
        & 0.371
        & 0.474
        & \second{0.369}
        & 0.370
        & 0.557
        & 0.377 \\

        & & 336
        & \best{0.383}
        & 0.408
        & 0.403
        & 0.395
        & 0.521
        & \second{0.390}
        & 0.399
        & 0.439
        & 0.403 \\

        & & 720
        & \best{0.413}
        & 0.443
        & 0.434
        & \second{0.425}
        & 0.529
        & \second{0.425}
        & 0.436
        & 0.438
        & 0.426 \\

        \cmidrule(lr){3-12}
        & & avg
        & \best{0.368}
        & 0.396
        & 0.390
        & 0.385
        & 0.487
        & \second{0.384}
        & 0.389
        & 0.452
        & 0.389 \\

        \midrule

        \multirow{10}{*}{ETTm2}
        & \multirow{5}{*}{MSE}
        & 96
        & \best{0.167}
        & 0.186
        & 0.174
        & 0.179
        & 0.212
        & 0.177
        & \second{0.171}
        & 0.225
        & 0.173 \\

        & & 192
        & \second{0.233}
        & 0.258
        & 0.240
        & 0.251
        & 0.274
        & 0.236
        & 0.235
        & 0.252
        & \best{0.229} \\

        & & 336
        & \second{0.282}
        & 0.317
        & 0.302
        & 0.329
        & 0.331
        & 0.290
        & \best{0.281}
        & 0.284
        & 0.288 \\

        & & 720
        & \best{0.372}
        & 0.393
        & 0.390
        & 0.479
        & 0.424
        & 0.388
        & \second{0.372}
        & 0.397
        & 0.377 \\

        \cmidrule(lr){3-12}
        & & avg
        & \best{0.264}
        & 0.289
        & 0.277
        & 0.310
        & 0.310
        & 0.273
        & \second{0.265}
        & 0.290
        & 0.267 \\

        \cmidrule(lr){2-12}

        & \multirow{5}{*}{MAE}
        & 96
        & \best{0.242}
        & 0.270
        & 0.259
        & 0.274
        & 0.282
        & 0.263
        & \second{0.253}
        & 0.319
        & 0.257 \\

        & & 192
        & \best{0.288}
        & 0.317
        & 0.303
        & 0.329
        & 0.324
        & 0.300
        & 0.299
        & 0.311
        & \second{0.294} \\

        & & 336
        & \best{0.325}
        & 0.354
        & 0.344
        & 0.385
        & 0.359
        & 0.337
        & \second{0.330}
        & 0.338
        & 0.332 \\

        & & 720
        & \best{0.381}
        & 0.401
        & 0.397
        & 0.477
        & 0.417
        & 0.393
        & \second{0.388}
        & 0.406
        & 0.389 \\

        \cmidrule(lr){3-12}
        & & avg
        & \best{0.309}
        & 0.336
        & 0.326
        & 0.366
        & 0.346
        & 0.323
        & \second{0.318}
        & 0.344
        & 0.318 \\

        \bottomrule
    \end{tabular}

    \caption{Full forecasting results on the ETT benchmarks.
    The best results are highlighted in \textcolor{red}{\textbf{red}},
    while the second-best results are
    \textcolor{blue}{\underline{blue}}.
    Rankings are determined using the original unrounded values.}
    \label{tab:ett_full_results}
\end{table*}
\begin{table*}[t]
    \centering
    \scriptsize
    \setlength{\tabcolsep}{3.2pt}
    \renewcommand{\arraystretch}{1.08}

    \begin{tabular}{ccc|ccccccccc}
        \toprule
        \textbf{Dataset}
        & \textbf{Metric}
        & \textbf{$P$}
        & \texttt{Ours}
& \texttt{iTransformer}
& \texttt{PatchTST}
& \texttt{DLinear}
& \texttt{TimesNet}
& \texttt{TiDE}
& \texttt{TimeMixer}
& \texttt{CNDiff}
& \texttt{TimeXer} \\
        \midrule

        \multirow{10}{*}{{Weather}}
        & \multirow{5}{*}{MSE}
        & 96
        & \best{0.152}
        & 0.166
        & 0.157
        & 0.183
        & 0.178
        & 0.187
        & 0.161
        & 0.261
        & \second{0.152} \\

        & & 192
        & \best{0.195}
        & 0.213
        & 0.205
        & 0.230
        & 0.240
        & 0.229
        & 0.201
        & 0.337
        & \second{0.197} \\

        & & 336
        & \best{0.248}
        & 0.263
        & 0.257
        & 0.270
        & 0.306
        & 0.280
        & 0.250
        & 0.263
        & \second{0.251} \\

        & & 720
        & \best{0.322}
        & 0.339
        & 0.335
        & 0.337
        & 0.403
        & 0.350
        & 0.325
        & 0.355
        & \second{0.326} \\

        \cmidrule(lr){3-12}
& & avg
& \best{0.229}
& 0.245
& 0.239
& 0.255
& 0.282
& 0.262
& 0.234
& 0.304
& \second{0.232} \\

        \cmidrule(lr){2-12}

        & \multirow{5}{*}{MAE}
        & 96
        & \best{0.185}
        & 0.210
        & 0.202
        & 0.242
        & 0.223
        & 0.230
        & 0.213
        & 0.295
        & \second{0.200} \\

        & & 192
        & \best{0.226}
        & 0.253
        & 0.246
        & 0.291
        & 0.277
        & 0.264
        & 0.246
        & 0.399
        & \second{0.245} \\

        & & 336
        & \best{0.272}
        & 0.291
        & \second{0.285}
        & 0.321
        & 0.317
        & 0.302
        & 0.287
        & 0.310
        & 0.287 \\

        & & 720
        & \best{0.333}
        & 0.343
        & 0.339
        & 0.377
        & 0.383
        & 0.348
        & 0.337
        & 0.367
        & \second{0.336} \\

        \cmidrule(lr){3-12}
& & avg
& \best{0.254}
& 0.274
& 0.268
& 0.308
& 0.300
& 0.286
& 0.271
& 0.343
& \second{0.267} \\

        \midrule

        \multirow{10}{*}{{Electricity}}
        & \multirow{5}{*}{MSE}
        & 96
        & 0.139
        & 0.137
        & 0.145
        & 0.152
        & 0.189
        & 0.155
        & \best{0.134}
        & 0.136
        & \second{0.135} \\

        & & 192
        & \second{0.154}
        & 0.157
        & 0.162
        & 0.164
        & 0.188
        & 0.167
        & \best{0.152}
        & 0.157
        & 0.158 \\

        & & 336
        & \best{0.171}
        & 0.175
        & 0.179
        & 0.183
        & 0.203
        & 0.193
        & 0.177
        & \second{0.174}
        & 0.175 \\

        & & 720
        & \second{0.207}
        & 0.221
        & 0.221
        & 0.217
        & 0.225
        & 0.233
        & 0.214
        & \best{0.200}
        & 0.217 \\

        \cmidrule(lr){3-12}
        & & avg
& \second{0.168}
& 0.173
& 0.177
& 0.179
& 0.201
& 0.187
& 0.169
& \best{0.167}
& 0.171 \\

        \cmidrule(lr){2-12}

        & \multirow{5}{*}{MAE}
        & 96
        & \best{0.222}
        & 0.231
        & 0.248
        & 0.247
        & 0.291
        & 0.249
        & \second{0.226}
        & 0.234
        & 0.232 \\

        & & 192
        & \best{0.236}
        & 0.251
        & 0.263
        & 0.259
        & 0.290
        & 0.259
        & \second{0.243}
        & 0.253
        & 0.258 \\

        & & 336
        & \best{0.253}
        & 0.267
        & 0.278
        & 0.282
        & 0.307
        & 0.288
        & \second{0.264}
        & 0.273
        & 0.270 \\

        & & 720
        & \best{0.286}
        & 0.307
        & 0.312
        & 0.314
        & 0.325
        & 0.319
        & \second{0.301}
        & \second{0.301}
        & 0.308 \\

        \cmidrule(lr){3-12}
        & & avg
& \best{0.249}
& 0.264
& 0.275
& 0.276
& 0.303
& 0.279
& \second{0.259}
& 0.265
& 0.267 \\

        \midrule

        \multirow{10}{*}{{Exchange}}
        & \multirow{5}{*}{MSE}
        & 96
        & \best{0.078}
        & 0.096
        & 0.091
        & 0.082
        & 0.164
        & 0.087
        & 0.086
        & \second{0.081}
        & 0.087 \\

        & & 192
        & \best{0.156}
        & 0.191
        & 0.186
        & \second{0.169}
        & 0.276
        & 0.180
        & 0.200
        & 0.175
        & 0.178 \\

        & & 336
        & \best{0.319}
        & 0.356
        & 0.405
        & \second{0.323}
        & 0.478
        & 0.334
        & 0.351
        & 0.351
        & 0.363 \\

        & & 720
        & \best{0.789}
        & 0.850
        & \second{0.848}
        & 0.918
        & 1.834
        & 0.852
        & 0.945
        & 0.978
        & 1.094 \\

        \cmidrule(lr){3-12}
        & & avg
& \best{0.336}
& 0.373
& 0.383
& 0.373
& 0.688
& \second{0.363}
& 0.396
& 0.396
& 0.431 \\

        \cmidrule(lr){2-12}

        & \multirow{5}{*}{MAE}
        & 96
        & \best{0.198}
        & 0.222
        & 0.212
        & 0.207
        & 0.268
        & 0.207
        & \second{0.205}
        & 0.197
        & 0.207 \\

        & & 192
        & \best{0.288}
        & 0.314
        & 0.307
        & 0.305
        & 0.378
        & 0.303
        & 0.321
        & \second{0.295}
        & 0.303 \\

        & & 336
        & \best{0.414}
        & 0.436
        & 0.462
        & 0.432
        & 0.501
        & \second{0.420}
        & 0.430
        & 0.428
        & 0.437 \\

        & & 720
        & \best{0.683}
        & 0.698
        & \second{0.693}
        & 0.721
        & 0.935
        & 0.695
        & 0.726
        & 0.729
        & 0.780 \\

        \cmidrule(lr){3-12}
        \cmidrule(lr){3-12}
& & avg
& \best{0.396}
& 0.418
& 0.419
& 0.416
& 0.521
& \second{0.406}
& 0.421
& 0.412
& 0.432 \\

        \midrule

        \multirow{10}{*}{{Traffic}}
        & \multirow{5}{*}{MSE}
        & 96
        & \best{0.397}
        & \second{0.398}
        & 0.409
        & 0.452
        & 0.597
        & 0.458
        & 0.405
        & 0.562
        & 0.406 \\

        & & 192
        & \best{0.416}
        & 0.426
        & \second{0.422}
        & 0.460
        & 0.619
        & 0.467
        & 0.429
        & 0.566
        & 0.425 \\

        & & 336
        & \best{0.424}
        & 0.441
        & 0.446
        & 0.485
        & 0.648
        & 0.518
        & \second{0.436}
        & 0.589
        & 0.445 \\

        & & 720
        & \best{0.454}
        & 0.491
        & 0.479
        & 0.515
        & 0.651
        & 0.520
        & \second{0.473}
        & 0.630
        & 0.482 \\

        \cmidrule(lr){3-12}
        & & avg
& \best{0.423}
& 0.439
& 0.439
& 0.478
& 0.629
& 0.491
& \second{0.436}
& 0.587
& 0.440 \\

        \cmidrule(lr){2-12}

        & \multirow{5}{*}{MAE}
        & 96
        & \best{0.242}
        & \second{0.277}
        & 0.281
        & 0.305
        & 0.325
        & 0.311
        & 0.284
        & 0.301
        & 0.279 \\

        & & 192
        & \best{0.247}
        & 0.297
        & \second{0.287}
        & 0.309
        & 0.334
        & 0.314
        & \second{0.287}
        & 0.302
        & \second{0.287} \\

        & & 336
        & \best{0.249}
        & 0.301
        & 0.300
        & 0.329
        & 0.337
        & 0.359
        & \second{0.297}
        & 0.311
        & 0.298 \\

        & & 720
        & \best{0.268}
        & 0.340
        & 0.320
        & 0.347
        & 0.352
        & 0.350
        & \second{0.314}
        & 0.322
        & 0.321 \\

        \cmidrule(lr){3-12}
        & & avg
& \best{0.252}
& 0.304
& 0.297
& 0.323
& 0.337
& 0.334
& \second{0.296}
& 0.309
& \second{0.296} \\

        \bottomrule
    \end{tabular}

    \caption{Full forecasting results on the remaining datasets. The best results are highlighted in
    \textcolor{red}{\textbf{red}}, while the second-best results are
    \textcolor{blue}{\underline{blue}}.}
    \label{tab:full_remain}
\end{table*}


\section{Reproducibility Statement}
\label{app:reproducibility}

All models are implemented in PyTorch and trained on a single A100
GPU with 40 GB of VRAM. 
We use 3 random seeds and report
mean of the performance.
Hyperparameters are selected using the validation set and remain fixed for test evaluation. 
We run all baseline results from their original repositories under our evaluation configuration.

\end{document}